\documentclass[pmlr]{jmlr}

\usepackage{booktabs}
\usepackage[load-configurations=version-1]{siunitx}

\theorembodyfont{\upshape}
\theoremheaderfont{\scshape}
\theorempostheader{:}
\theoremsep{\newline}

\usepackage{hyperref}

\let\svthefootnote\thefootnote
\newcommand\freefootnote[1]{%
  \let\thefootnote\relax%
  \footnotetext{#1}%
  \let\thefootnote\svthefootnote%
}
\usepackage{longtable}
\usepackage{afterpage}

\usepackage{array}
\newcolumntype{P}[1]{>{\centering\arraybackslash}p{#1}}
\usepackage{changepage}
\usepackage{bbding}
\usepackage{pifont}
\usepackage{stackengine,amssymb,graphicx,scalerel}

\jmlrvolume{1}
\jmlryear{2022}
\jmlrworkshop{DYAD'21 Workshop}

\title[Human Pose Forecasting in Face-to-Face Interaction Scenarios]{Comparison of Spatio-Temporal Models for Human Motion and Pose Forecasting in Face-to-Face Interaction Scenarios} 

   \author{\Name{German Barquero} \Email{gbarquga9@alumnes.ub.edu}\\
   \Name{Johnny Núñez} \Email{jnunezca11@alumnes.ub.edu}\\
     \addr Universitat de Barcelona and Computer Vision Center, Spain
     \AND
   \Name{Zhen Xu} \Email{xuzhen@4paradigm.com}\\
     \addr 4Paradigm, Beijing, China
     \AND
   \Name{Sergio Escalera} \Email{sergio@maia.ub.es}\\
     \addr Universitat de Barcelona and Computer Vision Center, Spain
     \AND
   \Name{Wei-Wei Tu} \Email{tuweiwei@4paradigm.com}\\
     \addr 4Paradigm, Beijing, China
     \AND
   \Name{Isabelle Guyon} \Email{guyon@chalearn.org}\\
     \addr LISN (CNRS/INRIA) Université Paris-Saclay, France, and ChaLearn, USA
     \AND
   \Name{Cristina Palmero} \Email{crpalmec7@alumnes.ub.edu}\\
     \addr Universitat de Barcelona and Computer Vision Center, Spain
     }

\begin{document}

\maketitle

\begin{abstract}
Human behavior forecasting during human-human interactions is of utmost importance to provide robotic or virtual agents with social intelligence. This problem is especially challenging for scenarios that are highly driven by interpersonal dynamics. In this work, we present the first systematic comparison of state-of-the-art approaches for behavior forecasting. To do so, we leverage whole-body annotations (face, body, and hands) from the very recently released UDIVA v0.5, which features face-to-face dyadic interactions. Our best attention-based approaches achieve state-of-the-art performance in UDIVA v0.5. We show that by autoregressively predicting the future with methods trained for the short-term future ($<$400ms), we outperform the baselines even for a considerably longer-term future (up to 2s). We also show that this finding holds when highly noisy annotations are used, which opens new horizons towards the use of weakly-supervised learning. Combined with large-scale datasets, this may help boost the advances in this field.
\end{abstract}

\section{Introduction}

Take a moment to think about a recent conversation you were engaged in. How often did you nod while endorsing the other interlocutor's speech? Or scratch your chin while thinking deeply about an answer to a difficult question? Both your compliance with someone's speech and a demanding question triggered specific behaviors. Now, imagine yourself involved in a conversation where your partner is completely frozen. Weird, right? These conversational behaviors are not only expected but also needed in order to enhance human-human communication. 

\begin{figure}[!t]
    \centering
    \includegraphics[width=\textwidth]{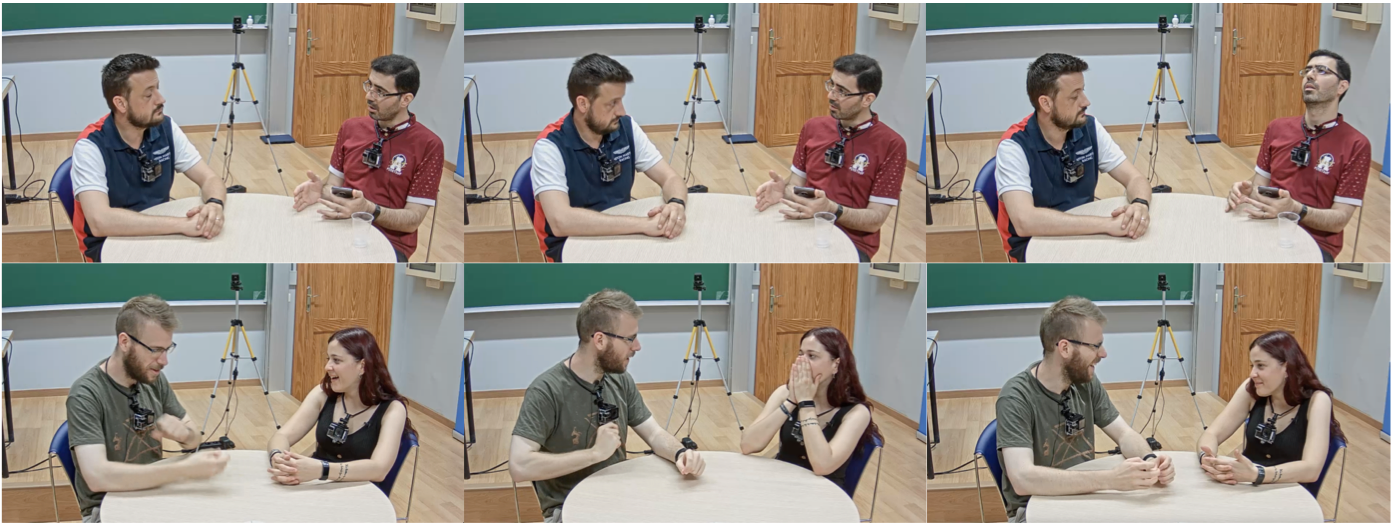}
    \caption{Two examples of conversations where each individual behavior is highly driven by the dyadic interaction. Top row: the left-most participant nods in answer to the interlocutor question, which results in the latter acquiring a typical posture of someone who is dubious and thinks deeply about something. Bottom row: both participants engage in a highly interactive conversation which triggers surprise and joy emotions from the right-most participant.}
    \label{fig:01_dyad_conver}
\end{figure}

The way someone behaves during a social interaction evolves over time, and can be driven by intrapersonal (e.g., personality, mood, culture), interpersonal (e.g., role within the interaction, shared history), and scenario (e.g., thesis defence, informal conversation in a restaurant) characteristics~\citep{reis2000relationship}. For example, the greeting between two people will depend on their relationship and their meeting context. The body language of an extrovert person talking about themselves will differ from that of a shy person. Similarly, someone's behavior will vary from a job interview to a friendly conversation. During these interactions, interlocutors can communicate and influence each other by means of verbal and non-verbal behaviors~\citep{burgoon2007interpersonal}, including facial expressions, gaze, gestures, or body language (see \autoref{fig:01_dyad_conver} for an example of interaction-driven behaviors). Even in contexts such as public speaking where the communication is mostly unidirectional, the speaker can perceive the public reaction by their subtle non-verbal cues and adapt his/her message and behavior accordingly.

Understanding the dynamics of individuals' behavior is important for many applications related to human communication~\citep{vinciarelli2009social, ondavs2019anticipation}. In medicine and psychology, for example, experts analyze the behavior of the patient in social environments in order to detect any communication anomalies. Deficits in eye contact, pointing or expressing emotions represent key markers that might help in the diagnosis of autism spectrum disorders~\citep{barbaro2013autism1, loth2018autism2}. In other fields like virtual reality or robotics, communication needs to be fully replicated. Therefore, social robots and virtual avatars need to combine all natural modalities (e.g., visual, audio, language) involved in a real interaction. In order to induce a human-like social interaction, their facial expressions~\citep{lombardi2018deep, moosaei2017using} need to be harmonized with the body pose, hands gestures and speech~\citep{saerbeck2010expressive}. They also need to tailor their behavior and communication style to the user and situation so as to increase perceived empathy and immediacy~\citep{marco2007face}. Most of these tasks can only be accomplished by understanding and modeling behavior in human-human interactions, which represents a challenge involving cognitive, affective, and behavioral perspectives, among others.

Another requirement especially important for human-robot communication or personalized assistive agents is the ability to forecast human behavior accurately and timely. Automated agents need to predict someone's short-term future behavior in order to provide a better and faster response, and a more natural communication~\citep{kanda2017human, nocentini2019survey}. Similarly, it is of utmost importance to any system to minimize the latencies. We cannot always afford that the user waits until the whole information is processed, so detecting and interpreting the user's social signals may help providing with a fast approximation as a starting point, which may be improved as more information is collected. For example, safety driving systems extremely benefit from anticipating pedestrian trajectories, as they may need to make an emergency stop in order to avoid an accident~\citep{alahi2016social}. 

One may argue that by forecasting behavior during a human-human interaction, we are also understanding the underlying mechanisms of human interaction and, therefore, uncovering their semantics, interdependencies, etc. Computational models for behavior forecasting have started to gain more popularity due to recent advances in computer vision, machine learning, and natural language processing techniques. In this sense, behavior can be expressed, represented, and predicted in many forms, ranging from low-level body landmarks or meshes to high-level action events and social signals~\citep{Barquero2022}. When using the former representation, this task is alternatively referred to as \textit{human motion prediction}~\citep{Corona2020} or \textit{human pose forecasting}~\citep{Ahuja2019}
. The most recent methodological advances in this task and the main challenges still to be addressed are discussed in the very recent survey from~\cite{Barquero2022}. Very recently, a behavior forecasting competition was held within the ChaLearn LAP DYAD@ICCV'21 challenge (hereafter referred to as DYAD'21), for which behavior was represented by means of body, face, and hand landmarks. The goal of the competition was to foster interlocutor- and context-aware models for social behavior understanding. It was based on UDIVA v0.5~\citep{Palmero2022}\footnote{\url{https://chalearnlap.cvc.uab.cat/dataset/41/description/}}, a dataset of face-to-face dyadic interaction sessions consisting of conversational, collaborative and competitive tasks in a lab setting with seated participants. The dataset includes videos and transcripts of the sessions, as well as participant (e.g., sociodemographics, self-reported personality, and pre- and post-session mood and fatigue), and session and dyadic metadata (e.g., characteristics of the tasks or relationship among interaction partners). In addition, it provides automatically extracted landmarks and 3D gaze vectors. None of the teams beat the competition baseline, a zero-velocity approach that propagated the landmarks of the last visible frame into the future. Nevertheless, the organizers identified some of the main challenges, such as the usage of noisy labels, the highly stochastic nature of the hands, or the mostly static nature of the dataset~\citep{Palmero2022}. Furthermore, human behavior is highly driven by a combination of short- and long-range temporal dependencies, which are difficult to capture appropriately~\citep{hausdorff1995walking}. 

\begin{figure}[!t]
    \centering
    \includegraphics[width=\textwidth]{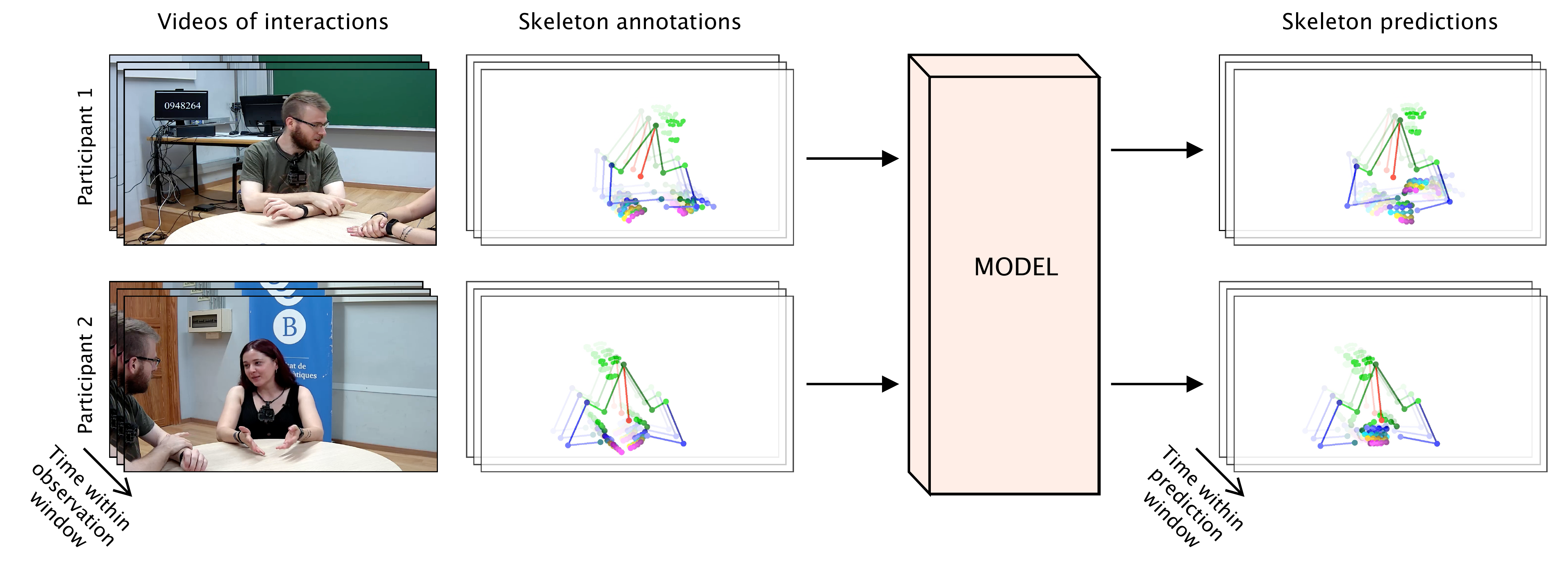}
    \caption{Illustration of the behavior forecasting problem tackled in this work. Given the automatically extracted skeleton landmarks from an \textit{observation window} in the past, the goal is to predict the skeleton motion within a \textit{prediction window} in the future.}
    \label{fig:problem_definition}
\end{figure}

Motivated by these findings, the goal of this work is to assess the performance of several state-of-the-art computer vision and machine learning approaches on the task of non-verbal social behavior forecasting, specifically represented as 2D face, body, and hands landmarks of the skeleton joints (hereafter referred to as \textit{skeleton}). In particular, we evaluate sequence-to-sequence (Seq2Seq) recurrent models~\citep{seq2seq2014}, Temporal Convolutional Networks (TCNs)~\citep{lea2017temporal}, Graph Convolutional Networks (GCNs)~\citep{stgnn2020}, and Transformers~\citep{vaswani2017attention}. All approaches are evaluated assuming that we are provided with a temporal window of observed behavior (\textit{observation window}) to predict the future behavior in a short time-frame (\textit{prediction window}) of up to 2 seconds. \autoref{fig:problem_definition} depicts the methodological workflow. Furthermore, we aim at exploring the effects of incorporating other available data modalities in addition to the visual modality, such as metadata about the participants in the dyad or transcripts. As previously stated, both intra- and interpersonal components influence human behavior and therefore need to be included in any behavioral model. However, contemplating all possible combinations of components yields a colossal set of scenarios. Additionally, behavior forecasting needs to deal with the stochasticity of the future. In order to alleviate both challenges, many related works focus on predicting the human behavior while performing specific actions \citep{diller2020forecasting, kania2021trajevae}. In these scenarios, there are no interpersonal cues and the intrapersonal ones are minimized. Also, the future is narrowed down and becomes more deterministic. Instead, we focus our work on dyadic conversations, where participants' behaviors are under the influence of strong interpersonal cues. To do so, our comparison is applied on the UDIVA v0.5 dataset. Apart from following the challenge's benchmark so that our results can be compared, the dataset choice came naturally for three main reasons. First, their diversity with respect to participants and contexts provides a reasonable playground for modeling behavior. Second, it is the only dataset up to date that provides landmarks for all the visible parts of the body, including face and hands, for dyadic social interactions~\citep{Barquero2022}. Lastly, the cleaned annotations for the validation and test sets include manual fixes for up to 7\% of the hands, and quality labels for each part of the body in every frame. This allows us to assess the influence that noise present in a real deployment would imply in the task of behavior forecasting (Section~\ref{sec:exp_robustness_noise}).

Our experimental evaluation showed that the models trained to generate short-term behavior ($<400ms$) can be autoregressively applied to predict longer-term behavior (up to 2 seconds) and outperform the baselines for the long term (Section~\ref{sec:exp_unimodal}). The robustness experiment proved the superiority of our models with respect to the baseline in a hypothetical real-life deployment with uncleaned noisy annotations (Section~\ref{sec:exp_robustness_noise}), especially for the short term future. The overall good results obtained in this work by training with noisy automatically extracted annotations proves the feasibility of weakly-supervised training for the behavior forecasting task in UDIVA v0.5. Finally, our experiments on multimodal and dyadic-aware models showed that more complex fusion strategies need to be explored in order to efficiently exploit multimodality and interpersonal dynamics.

To the best of our knowledge, this is the first systematic comparison of state-of-the-art approaches for behavior forecasting~\citep{Barquero2022}, and also the first in extensively evaluating the potential of UDIVA v0.5 for such task. The main contributions of this work are summarized as follows:
\begin{itemize}
    
    \item We present several adaptations of recurrent, convolutional, attention-based, and graph networks for behavior forecasting. We propose a Transformer-based method which becomes the state of the art for behavior forecasting with UDIVA v0.5\footnote{The code is publicly available in \url{https://github.com/crisie/UDIVA/tree/main/BehaviorForecasting}}. Its promising results even when trained with highly noisy annotations open new horizons towards the use of weakly-supervised learning for behavior forecasting.
    
    \item We analyze the main challenges of behavior forecasting specific to a conversational scenario. In particular, its static nature, its high percentage of noise, and the highly dyad-driven behavioral dynamics make it a very challenging benchmark.
    
    \item We present the results of several attempts to model the interpersonal dynamics which drive a dyadic conversation. They serve us to state the current limitations of the dataset, and how future work may try to address them.
    
\end{itemize}

The remainder of this paper is organized as follows. Section~\ref{sec:method} describes the methodology followed for the systematic comparison, including details of different modalities of input data used, data processing, and models compared. In Section~\ref{sec:experimental_evaluation}, we present the details of the experimental evaluation carried out. Their descriptions are complemented in Section~\ref{sec:discussion} with a thorough discussion of the findings of our work, the main challenges posed by UDIVA v0.5, and the most relevant future research lines in this field. Finally, in Section~\ref{sec:ethics}, we review the ethical considerations regarding human motion and pose forecasting and its real-world applications.

\section{Methodologies}
\label{sec:method}

We frame our methodological approach within the non-verbal behavior forecasting taxonomy of~\citet{Barquero2022}. 
In particular, we aim at predicting \textit{low-level} behavioral representations: face, body, and hands 2D landmarks. The motivation behind using the visible skeleton
is delineated with our goal of modeling as many dimensions of the human behavior as we can. In fact, we might consider our problem to be an implicit generalization of the forecasting of any visual social cue or signal that can be perceived from facial, body, or hands landmarks. 

Even though such problem formulation could be seen as a multi-task problem, in which each part of the body is predicted separately, we see the human skeleton as a single skeleton and therefore consider our framework to be defined in a \textit{single task} scenario. Further, we conceive our approach under the \textit{deterministic} viewpoint. That is, both our methods and evaluation metrics consider a single possible future. We also explore the benefits and challenges of modeling the dyad-driven dynamics by evaluating three \textit{context-aware} methodologies. In our scenario, the context is made up of the pose, motion, and any information (e.g., speech, metadata) belonging to the interactant. Additionally, we explore the use of several \textit{multimodal} combinations of data including text, audio, and metadata features. Finally, although we use features from observed time windows that are longer than in most related works, we still consider us to be on the \textit{history-blinded} side. The reason is that we are not implementing any specific strategy to extract insights from the historical data.

In Section~\ref{sec:data_preparation}, we describe the pre-processing steps for all the data used from the dataset: skeleton annotations, metadata, audio, and transcriptions. Then, we present the common encoder-decoder framework shared among all the methods tested (Section~\ref{sec:archs}).

\subsection{Data preparation}
\label{sec:data_preparation}

\subsubsection{Pose representation}

In first place, it is important to note that, since the UDIVA skeleton annotations were automatically extracted and post-processed, they are highly noisy (around 20\% of frames had wrong hand annotations)~\citep{Palmero2022}. On the positive side, this allowed us to conceptually train our models in a weakly-supervised fashion. At the same time though, the frequent wrong or missing landmark estimations posed new challenges to the training stability and model performance.

Additionally, there were several challenges associated to the skeleton annotations that needed to be addressed. In the first place, the three skeleton entities (face, body and hands) were independently extracted. As a result, the depth coordinates among them did not match. This made it more difficult for the model to extract useful insight from the 3D coordinates. Secondly, our predictions needed to be detached from the image space in order to compress the subspace of poses that the network needed to model. At the same time, the location and trajectory of the person with respect to the camera view (e.g., centered, closer to its interlocutor, etc.) was relevant and strongly determined their future behavior. Therefore, our predictions could not be fully trajectory-agnostic.

The first challenge was unsolvable, so we assumed that our three spaces of coordinates would be independent and that the depth coordinate could become an important source of noise. In order to alleviate the second problem, some works proposed working with offsets instead of poses. This strategy isolates the input feature from the image space, and thus narrows their range of possible values~\citep{Martinez2017, Adeli2020}. Unfortunately, this solution completely removed the information about the global position of the participants, which was not favorable either. In order to find the right trade-off, recent works have shown that mixing root-relative position and velocity was beneficial for future single human motion forecasting~\citep{liu2020deepssm, li2021symbiotic}. Following this trend, we combined the joints' root-relative coordinates and their offsets to create our input. Our root points were chosen as 1) the middle point between both eye centers for face, 2) the middle chest joint for body, and 3) the middle knuckles for hands. 
Therefore, our input skeleton vectors consisted of root-relative 3D coordinates and 3D offsets for 28 inner facial landmarks, 10 upper-body joints, and 40 hand landmarks (20 for each hand)~\citep{Palmero2022}, which can be visualized in \autoref{fig:problem_definition}. Missing sets of landmarks for face, body, or hands were replaced by sets of zero-valued landmarks. Additionally, the global image coordinates and offsets of the face, body, and hands roots were also included in the skeleton input. Finally, the gaze vector directions and its offsets were also included as input features (although gaze vector directions were not predicted).

\subsubsection{Multimodal data}
\label{sec:inputs_multimodal}

Our work includes a first attempt to leverage multimodal data to forecast low-level representations of future behavior. The metadata, transcriptions, and the audio were processed so that they could be easily fed into any network.

\textbf{Metadata.} For each session, we generated a vector of normalized \textit{participant} and \textit{session} metadata as proposed in \cite{palmero2020context}. Additionally, we used education, and the speaking language. In particular, the age (scaled from [17,75] to [0,1]), gender (male, 0; female, 1), country of origin (6D one-hot encoding after recategorization based on cultural differences, \citealt{mensah2013global}), education level (7D one-hot encoding) and self-reported personality (z-scores of the Big Five traits, \citealt{mccrae1992introduction}) were included as participants' metadata. The session metadata comprised the language spoken (3D one-hot encoding for English, Spanish, and Catalan), the relationship between participants (unknown, 0; known, 1), the pre-session mood (8D vector, scaled from [1,5] to [0,1] each) and the pre-session fatigue scores (scaled from [0, 10] to [0,1]). The mood was assessed with 8 categorical values: good, bad, happy, sad, friendly, unfriendly, tense, and relaxed.

\textbf{Audio.} The audio features were extracted with the pre-trained VGGish model~\citep{hershey2017vggish}. For each frame $T$, the model generated a feature vector of size 128 with the audio chunk of 1 second (25 frames) centered at $T$. All audio features were normalized to the [-1, 1] range with statistics generated from the training sessions. Note that the audio feature vectors for the last 12 frames of the observation window were set to zero because their generation used audio that overlapped the prediction window.

\textbf{Transcriptions.} To avoid an explosion of dimensionality driven by the wide spectrum of speech possibilities, and in order to standardize features retrieval across observation windows, we did not use utterance-based embeddings. Instead, for each prediction window, all the utterances that overlapped its preceding observation window and ended on it were jointly considered. A state-of-the-art sentence encoder network was used to extract semantic embeddings of size 768 from the combined utterances \citep{reimers-2019-sentence-bert}.

\subsection{Settings tested}
\label{sec:archs}
The architectures included in our comparison have been built over an encoder-decoder architecture and adapted to the needs of each of the settings explored. Each setting is defined by the origin of the information leveraged: from 1) a single participant and a single modality (unimodal), 2) both participants and a single modality (dyadic), and 3) a single participant and two modalities (multimodal).

\begin{figure}[!t]
    \centering
    \includegraphics[width=\textwidth]{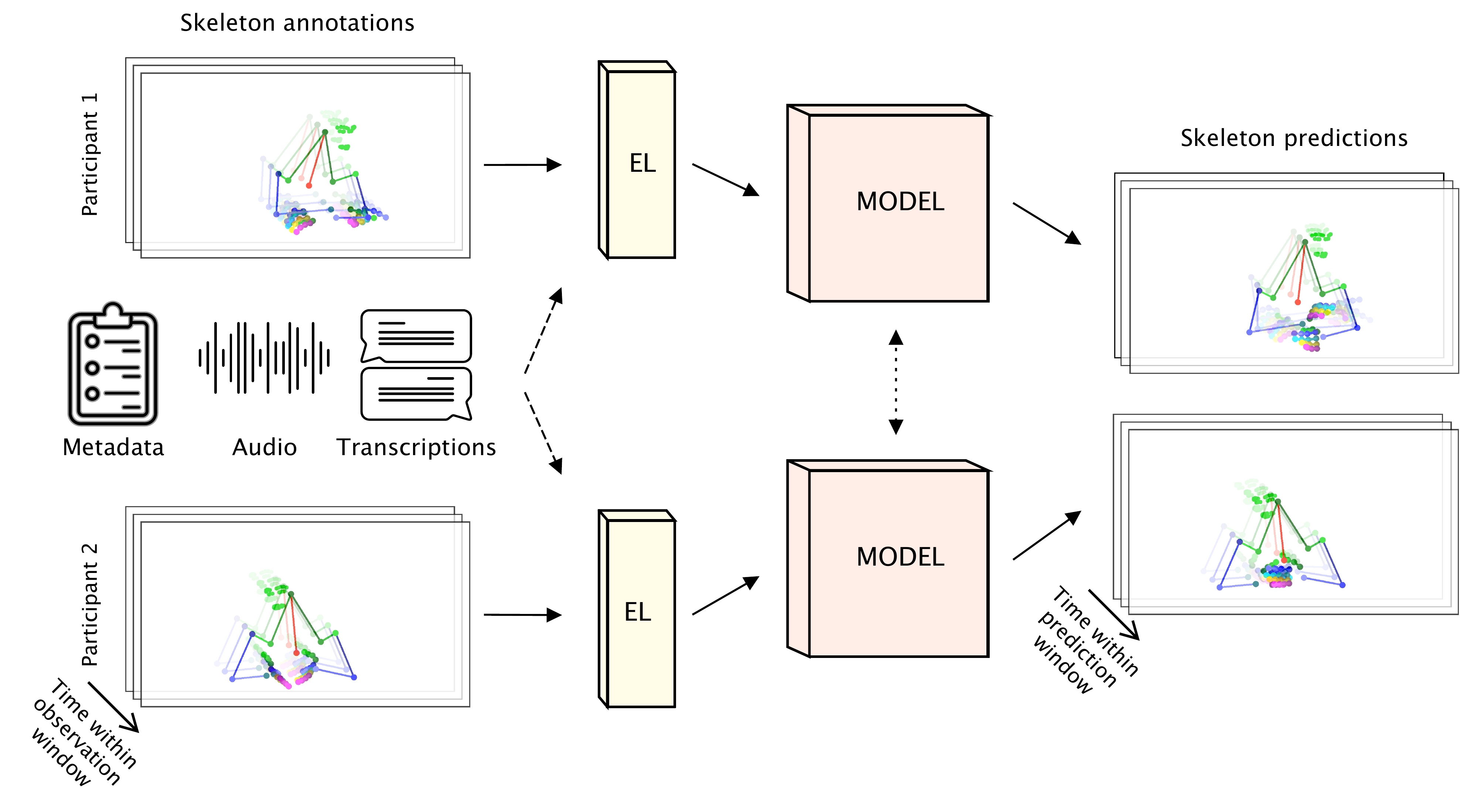}
    \caption{Overview of the modular encoder-decoder architecture used in the experiments. The embedding layer (EL) uses a dense layer to generate an intermediate representation of the coordinates and the offsets from the observed skeletons. The dashed and the dotted arrows are active only for the multimodal and the dyadic experiments, respectively. The output of the model are always 2D offsets, either implicitly by the implementation of a residual layer (recurrent decoding), or explicitly (non-recurrent decoding). Weights for both models are shared.}
    \label{fig:architectures_overview}
\end{figure}

\subsubsection{Unimodal}
\label{sec:arch_unimodal}

We built an encoder-decoder architecture to test several encoding and decoding state-of-the-art modules. As shown in our architecture scheme (see \autoref{fig:architectures_overview}), before going through the model, the observed skeletons are fed to an embedding layer (EL) that combines the landmarks coordinates and offsets. The models used for both participants share their weights. The technical details about the four architectures tested are included in the Appendix~\ref{app:arch_details}, and briefly described next:

\begin{enumerate}
    \item Seq2Seq. Adopted from the natural language processing field~\citep{seq2seq2014}, the sequence-to-sequence model has always been a very popular framework for human motion forecasting~\citep{Martinez2017, li2018convolutional}. We test the Seq2Seq with a single gated recurrent unit, Seq2Seq w/ GRU~\citep{gru2014}, or a single long short-term memory unit, Seq2Seq w/ LSTM~\citep{lstm1997}.
    
    \item TCN. Temporal convolutional networks, TCNs~\citep{lea2017temporal}, have proven to work as well as recurrent models in many practical applications~\citep{tcn2018}. Their main advantage resides in their ability to model long-range dynamics with relatively few parameters. In our work, we use it to encode the input. The decoding is done by a single GRU/LSTM unit. For the former, the encoded vector is used as the hidden state. For the latter, the encoded vector is split into halves to generate the initial hidden and cell states.
    
    \item STGNN. In spatiotemporal multivariate time series modeling, Spatial Temporal Graph Neural Networks (STGNN) has been a popular choice. The STGNN model proposed by \cite{stgnn2020} has shown to be state-of-the-art. The model consists of a series of temporal module and spatial module. Temporal module is implemented with dilated TCN while spatial module is implemented with mixhop graph neural networks. In graph convolution, the graph structure between nodes are learnt end-to-end thus no ground truth adjacency matrix is needed. 
    
    \item Transformer. Also originally from the natural language processing world, Transformers are attention-based architectures that are populating the top positions in many tasks' benchmarks. In this work, we test the spatio-temporal transformer defined by \cite{poseformer2021} (Transformer\_ST). A simplification without spatial attention is also considered (Transformer\_T). Both architectures use a single GRU unit as decoder.
\end{enumerate}

The number of parameters of the TCN and STGNN models is kept about 10 times lower than Seq2Seq and Transformer models in order to avoid intractable training times. All recurrent decoders implement a residual connection between the $x$ and $y$ coordinates of the recurrent unit input and the output, so that the network learns to predict 2D offsets instead of 2D coordinates. For STGNN, we add up the prediction frames to the last observation to force the model predict offsets.

\subsubsection{Dyadic}
\label{sec:arch_dyadic}

We adapted the Seq2Seq, TCN, and Transformer models so that they received the landmarks from both participants. Our attempts so that the network effectively models the dyad-driven aspects of the interaction include:

\textbf{Early fusion.} We first explore the simplest way: modeling the behavior of both participants jointly since the beginning. Therefore, the observed skeletons of both participants are concatenated immediately after the embedding layer, and encoded together. The decoding stage is equivalent to that of the unimodal models.

\textbf{Late fusion.} In our second attempt, each participant is independently encoded. Then, before every decoding step, the input embedded skeleton is concatenated with the encoded vector of the other participant, which is constant during the whole decoding process.

\textbf{Interactive.} Additionally, we implemented the fusion method proposed by \cite{Honda2020}. Unfortunately, the encoding part could not be implemented exactly as described for TCN and Transformers due to the absence of a recurrently updated hidden state. In these cases, the encoder simply applies early fusion instead. The decoding part is interactive in the sense that the evolving hidden states are exchanged between decoders at each time step. The main difference with respect to the late fusion strategy is that the interactive one is aware of the future predictions made in parallel for the other participant. As a result, both participants' predictions should maintain certain coherence.

Note that, during training and testing stages of all the dyadic models presented, the prediction is done simultaneously for both participants. Therefore, in the late fusion approach, the encoded vector of participant B that is fed with the embedded skeleton into the decoder of participant A is the same that the decoder of participant B uses to carry out its own decoding. Similarly for the interactive approach.

\subsubsection{Multimodal}
\label{sec:arch_multimodal}

As a first attempt to leverage multimodal information for behavior forecasting, we explored the inclusion of the metadata, transcriptions, and audio of the target participant (monadic). In preliminary unimodal experiments (see Section~\ref{sec:exp_unimodal}), transformer-based methods outperformed the others, so for simplicity we built our multimodal approaches with Transformer\_T:

\textbf{Metadata.} Intuitively, values such as the participant tiredness or mood before starting the session are important for both modeling a more accurate human behavior into the latent space (encoder), and also inferring an appropriate behavior continuation (decoder). Therefore, the target participant's metadata array is embedded by a dense layer and then concatenated to the embedded skeleton of both the encoder and decoder in every step. Given the impact that personalizing behavior forecasting to each personality idiosyncrasies may have, especially in the long term, we consider two configurations for the metadata array: with and without personality traits. Such metadata array consists of 34 values when personality traits are considered (29 if not), and does not include the interactant's mood, fatigue, or demographic metadata.

\textbf{Transcriptions.} Similarly, the transcriptions are also embedded with a dense layer and concatenated with the embedded skeleton at every encoding and decoding steps.

\textbf{Audio.} Differently from previous modalities, the audio features are extracted and embedded through a dense layer in a frame basis. The audio and skeleton embeddings are concatenated for every frame of the observation window. In contrast with the previous multimodal approaches, the decoder is equivalent to that of the unimodal approach. This decision was motivated by the lack of a single audio feature representative of the whole observation window. Although we could have simply selected the latest audio feature vector, or embedded all of them into a single representation, we wanted to promote simplicity. In any case, the most relevant audio features can flow through the decoder by means of the encoding process. 

\section{Experimental evaluation}
\label{sec:experimental_evaluation}

In this section, we first present the baselines, the training details and the evaluation protocol (Sections~\ref{sec:baselines}, \ref{sec:exp_training} and \ref{sec:evaluation_metrics}, respectively). In our first experiment, we describe the effects of training with short- and long-term losses (Section~\ref{sec:exp_unimodal}). We follow by exploring whether building a single model specific for the motion prediction of each part of the body is a better choice than the holistic approach (Section~\ref{sec:exp_holistic}). Next, we show the performance of the dyadic-aware and multimodal approaches (Sections~\ref{sec:exp_dyadic} and \ref{sec:exp_multimodal}). Finally, we present three experiments that go deeper in the analysis of more specific issues like the inference with noisy labels (Section~\ref{sec:exp_robustness_noise}), the appropriate length of the observation window (Section~\ref{sec:exp_attention}), and the performance of our best models at the DYAD'21 behavior forecasting competition (Section~\ref{sec:challenge}). All the findings from the experiments are further discussed in Section~\ref{sec:discussion}.

\subsection{Baselines}
\label{sec:baselines}
First, in order to make sure that our models are doing better than a naive approach, we establish three simple but challenging metrics.

\textbf{Zero-velocity.} The simplest baseline for behavior forecasting using skeletons consists in propagating the landmarks from the last observed frame into the future, as if the person froze once the observation window finished (\textit{zero-velocity}). While it may seem counterintuitive, the zero-velocity baseline has been proven to be a very strong and difficult to improve baseline~\citep{Martinez2017}. This is especially true in our conversational validation and test sets, as most parts of the participant's body remain static while listening to their partner's speech. In fact, it was the DYAD'21 competition baseline.

\textbf{Linear propagation.} Inspired by \cite{Adeli2020}, this baseline predicts the global motion for each part of the body as the linear propagation of its last observed velocity~(\textit{LinearProp}).
This velocity is computed as the average velocity of all landmarks from that part of the body. 

\textbf{Regression to observed mean.} This baseline is designed to replicate the regression-to-the-mean effect commonly observed in forecasting problems like ours~\citep{Feng2017, Raman2021}. To do so, it computes the linear velocity needed to get from the last observed skeleton pose to the mean observed skeleton pose in 2 seconds (at the last predicted frame). 
Two versions of this baseline were considered. The first (\textit{RTOMean}) uses the same constant velocity model for all the landmarks from each part of the body (face, body, and hands). The second (\textit{RTOMean\_L}) makes the regression with one constant velocity model per landmark. Intuitively, while the first simply translates the last observed face, body and hands to their observed mean positions, the second also rotates and scales each part of the body to their observed mean shape.

\subsection{Training details}
\label{sec:exp_training}

Our goal is to learn to predict behavior for the \textit{Talk} task. However, the other tasks provided in UDIVA v0.5 (\textit{Animals}, \textit{Lego}, and \textit{Ghost}) also contain conversational parts that share many behavioral dimensions with the target task. Therefore, we used the raw annotations of all the tasks (\textit{Talk}, \textit{Animals}, \textit{Lego}, and \textit{Ghost}) from the training sessions of UDIVA v0.5 as training set in order to include as much diverse behaviors as possible. For validation, though, only the cleaned annotations from the \textit{Talk} task of the UDIVA v0.5 validation sessions were used. Training and validation segments were extracted with a stride of 50 frames. Note that the validation segments are therefore different from those provided in the DYAD'21 challenge, which were manually selected by the organizers. Unless otherwise specified, an observation window of 100 frames (4 seconds) was used. The prediction window was fixed to 50 frames (2 seconds) in order to match the formulation of the challenge. The training loss used was the mean-squared error. 
All the models were trained with the AMSGrad variant of the Adam optimizer \citep{kingma2014adam} (weight decay to 0.001, $\beta_1$=0.9, $\beta_2$=0.999) with learning rate set to 0.0001. The batch size and dropout values were tuned for each model type. The final values after hyperparameter optimization are: 512 and 0.5 for Seq2Seq, 512 and 0.25 for TCN, 32 and 0.25 for Transformer, and 32 and 0.3 for STGNN. Early stopping was applied to all models with a patience of 20 epochs.

In addition, in order to deal with the particularities of the dataset, its associated metrics, and the problem itself, several training decisions were taken:

\begin{enumerate}
    \item Missing landmark estimations and hidden parts of the body (e.g., hands under the table) were fed to the network as zeros. As learning to predict such hidden states was beyond the scope of this work, hidden or missing landmarks were filtered out when computing the training loss (similar strategy as in \citealt{Adeli2021}).
    
    \item From both training and validation sets, segments with a missing hand in the last observed frame but visible in the sequence to predict were filtered out (21.5\% and 6.0\% of segments, respectively). The rationale behind this decision is purely conceptual. Since the network implements a residual connection and therefore predicts the offsets between consecutive frames, the full skeleton of the predicted sequence cannot be reconstructed if the hand is missing in the last frame of the observation window.
    
\end{enumerate}

\subsection{Evaluation protocol}
\label{sec:evaluation_metrics}

The evaluation was performed on the cleaned annotations of the UDIVA v0.5 test set (11 sessions), which include a selection of 278 prediction windows of 2 seconds from the \textit{Talk} task. 
Our models are evaluated with the Mean Per Joint Position Error (MPJPE), which has been broadly used in prior behavior forecasting works~\citep{Adeli2021, Wang2021Multi}. It is computed as the mean of the L2 distances between the ground truth and the predicted landmarks in all frames.
We also propose the short-term (ST), the mid-term (MT), and the long-term (LT) variants, whose computation is restricted to the subintervals of frames 1-10 (0-400ms), 11-25 (400ms-1s) and 26-50 (1s-2s), respectively. We also calculate the Final
Displacement Error (FDE,~\citealt{mohamed2020social}), which corresponds to the MPJPE of the last predicted frame of the sequence, and helps to detect error propagations to the very long-term. Finally, a divergence metric ($\Delta$) is calculated as the mean displacement per frame and landmark (pixels) in the predicted sequence. This metric helps to quantify the amount of motion present in the predicted sequences of skeletons. Similarly to MPJPE, the short-, mid-, and long-term versions are also present.

As reported in \cite{Palmero2022}, cleaned annotations of UDIVA v0.5 underwent visual inspection. Despite all wrong landmark extractions were identified, only a subset of them were fixed. Therefore, it is important to note that our observed skeletons, even for evaluation, are noisy per se (about 15\%, 3\%, 1\% of inaccurate hands, body, and face observations, respectively). Nonetheless, all evaluation metrics are computed only on the correct landmarks. Missing face, body, or hands annotations are also left out.

Additionally, and in order to compare our methods to those from the DYAD'21 challenge, we consider the face, body, and hands metrics presented in \cite{Palmero2022}, and present them in Section~\ref{sec:challenge}.




\subsection{Unimodal results}
\label{sec:exp_unimodal}

\addtolength{\tabcolsep}{-2pt} 
\begin{table}[t!]
    \footnotesize
    \centering
    \begin{tabular}{lc@{\hskip 3mm}ccccc@{\hskip 6mm}cccc}
    \toprule
    & Params. & MPJPE & ST & MT & LT & FDE & $\Delta$ & $\Delta_{ST}$ & $\Delta_{MT}$ & $\Delta_{LT}$ \\
    \midrule
Ground truth & - & - & - & - & - & - & 0.86 & 0.82 & 0.85 & 0.88 \\
\midrule
\multicolumn{11}{c}{Baseline models} \\
\midrule
Zero-velocity & - & 16.16 & 6.00 & 14.45 & 21.12 & 23.37 & 0.00 & 0.00 & 0.00 & 0.00 \\
LinearProp & - & 34.88 & 7.65 & 25.84 & 51.06 & 65.22 & 0.64 & 0.64 & 0.64 & 0.64 \\
RTOMean & - & 17.60 & 6.46 & 15.53 & 23.15 & 25.73 & 0.19 & 0.19 & 0.19 & 0.19 \\
RTOMean\_L & - & 17.60 & 6.48 & 15.53 & 23.14 & 25.72 & 0.21 & 0.21 & 0.21 & 0.21 \\
\midrule
\multicolumn{11}{c}{Long-term models} \\
\midrule
Seq2Seq w/ LSTM & 15.2M & 16.46 & 6.03 & 14.66 & 21.58 & 24.08 & 0.12 & 0.08 & 0.12 & 0.13 \\
Seq2Seq w/ GRU & 12.0M & 16.74 & 6.03 & 14.76 & 22.07 & 24.68 & 0.14 & 0.12 & 0.15 & 0.14 \\
TCN + LSTM & 1.5M & 17.25 & \textbf{5.89} & 14.88 & 23.09 & 26.26 & 0.16 & 0.20 & 0.16 & 0.15 \\
TCN + GRU & 1.7M & 16.95 & 6.06 & 14.91 & 22.39 & 25.20 & 0.15 & 0.19 & 0.15 & 0.12 \\
STGNN & 1.5M & \textbf{16.42} & 5.93 & 14.78 & \textbf{21.48} & \textbf{23.69} & 0.28 & 0.44 & 0.34 & 0.19 \\
Transformer\_T & 15.3M & 17.27 & 6.24 & 15.30 & 22.72 & 25.38 & 0.17 & 0.25 & 0.17 & 0.13 \\
Transformer\_ST & 15.0M & 16.81 & 5.99 & 14.84 & 22.18 & 24.72 & 0.15 & 0.21 & 0.15 & 0.12 \\
\midrule
\multicolumn{11}{c}{Short-term models} \\
\midrule
Seq2Seq w/ LSTM & 15.2M & 15.86 & 5.75 & 14.32 & 20.69 & 22.78 & 0.07 & 0.18 & 0.05 & 0.04 \\
Seq2Seq w/ GRU & 12.0M & 15.96 & 5.73 & 14.36 & 20.88 & 23.18 & 0.09 & 0.21 & 0.06 & 0.06 \\
TCN + LSTM & 1.5M & 16.12 & 5.53 & 14.13 & 21.41 & 24.73 & 0.14 & 0.23 & 0.07 & 0.16 \\
TCN + GRU & 1.7M & 16.28 & 5.45 & 14.07 & 21.79 & 25.59 & 0.17 & 0.25 & 0.09 & 0.20 \\
STGNN & 1.5M & 17.17 & 5.41 & 14.45 & 23.37 & 27.50 & 0.23 & 0.29 & 0.16 & 0.25 \\
Transformer\_T & 15.3M & 15.65 & 5.44 & \textbf{13.93} & 20.63 & 23.06 & 0.12 & 0.29 & 0.08 & 0.08 \\
Transformer\_ST & 15.0M & \textbf{15.62} & \textbf{5.34} & 13.95 & \textbf{20.59} & \textbf{22.74} & 0.09 & 0.27 & 0.07 & 0.04 \\
 \bottomrule
    \end{tabular}
    \caption{Comparison of models optimized considering the whole prediction window of 2s (long-term models) and those optimized with respect to only 400ms (short-term models). Lower values for MPJPE and FDE are better. Abbreviations: MPJPE, mean per joint position error; ST/MT/LT, short-/mid-/long-term mean per joint position error; FDE, final displacement error; $\Delta$, divergence; $\Delta_{ST/MT/LT}$, short-/mid-/long-term divergence.}
    \label{tab:unimodal_comparison}
\end{table}

All unimodal models were trained with two configurations:
\begin{itemize}
    \item \textit{Short-term methods.} The number of frames that networks outputted was 10, for which the mean-squared error loss was computed and backpropagated. The validation loss was also computed considering only 10 frames. Longer-term predictions (up to 50 frames) were autoregressively generated at inference time by including the 10 predicted skeletons in the next observation window. That is, the prediction windows $[t, t+10)$ and $[t+10, t+20)$ were generated using $[t-100, t)$ and $[t-90, t+10)$ as observation windows, respectively.
    \item \textit{Long-term methods.} The number of frames that networks outputted was 50. The training loss was computed and backpropagated for all of them.
\end{itemize}

Results are shown in \autoref{tab:unimodal_comparison}. Additionally, metrics restricted to each part of the body are shown in Tables~\ref{tab:st_face_results}, \ref{tab:st_body_results}, and \ref{tab:st_hands_results} in Appendix~\ref{app:unimodal_split}.
Interestingly, the zero-velocity baseline outperforms all other baselines. As we expected, the LinearProp shows poor performance due to the high error originated in segments with fast motion in the last observed frames. Both versions of regression to the observed mean provide a good non-static metrics' upper bound. None of the long-term methods outperform the zero-velocity baseline in the overall MPJPE. Among them, the STGNN seems to perform the best. For all models with the long-term loss, we consistently observed that the validation loss quickly reached the validation minimum and started diverging (overfitting). Very interestingly, most short-term approaches outperform not only their long-term versions but also the baseline results in the long-term horizon. This is specially striking because they were not specifically trained to make long-term predictions. In contrast to the long-term experiment, the validation loss for the short-term models decreased steadier, taking considerably longer to overfit. Among the short-term architectures, both transformer-based approaches performed the best in all time-windows. The divergence values show a common freezing trend for all short-term methods except for the STGNN as they go deep into the future. On the contrary, $\Delta_{LT}$ increases for short-term TCN methods, suggesting their higher sensitivity to noise.

\begin{figure}
    \centering
    \includegraphics[width=\textwidth]{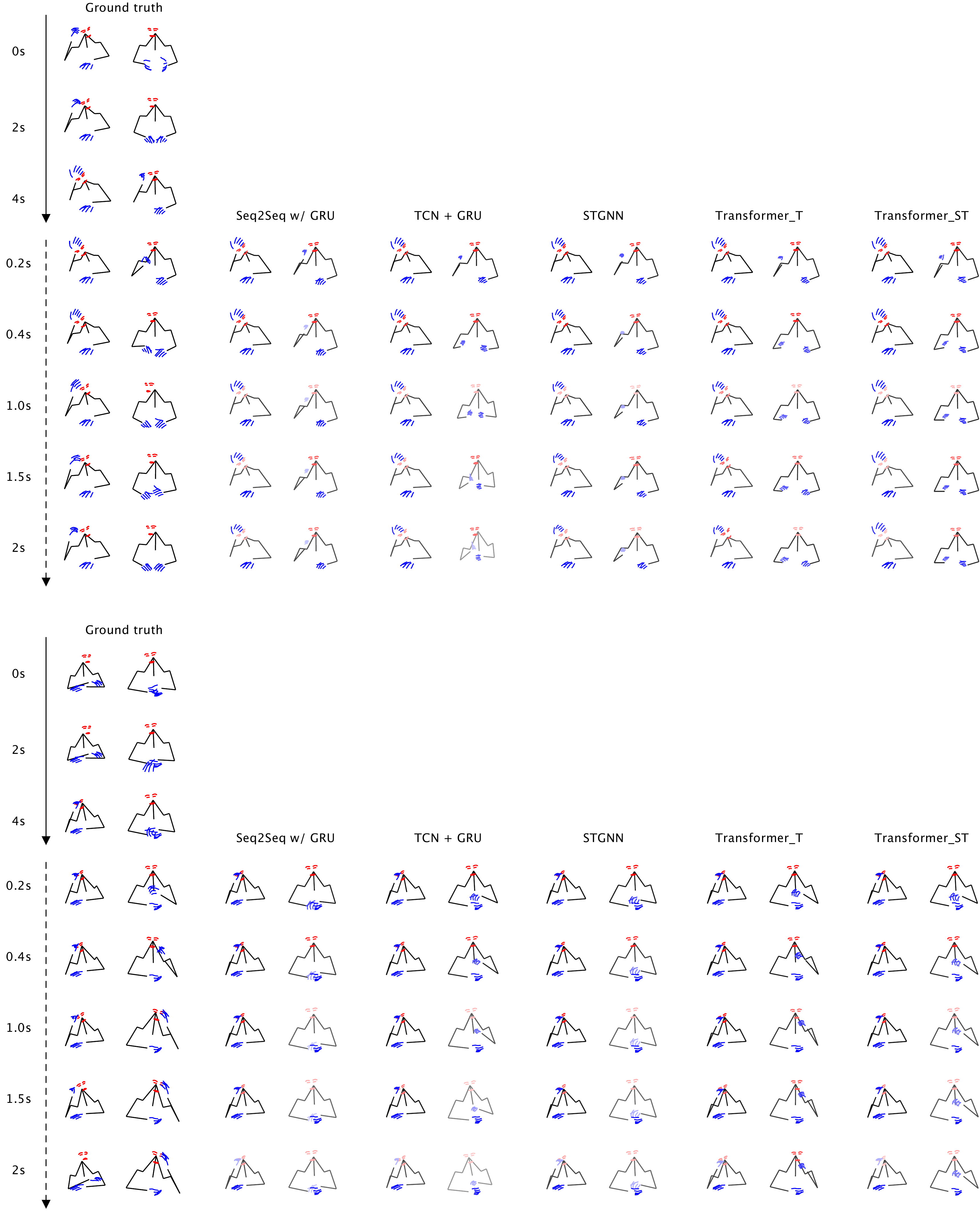}
    \caption{Two examples of behavior forecasting by the tested architectures. The solid and dashed arrows correspond to the observed and predicted temporal windows, respectively. The higher the transparency, the higher the error for that part of the body is. In both cases, a high variance among the sequences predicted by the models is observed.}
    \label{fig:visualization}
\end{figure}

\begin{figure}
    \centering
    \includegraphics[width=\textwidth]{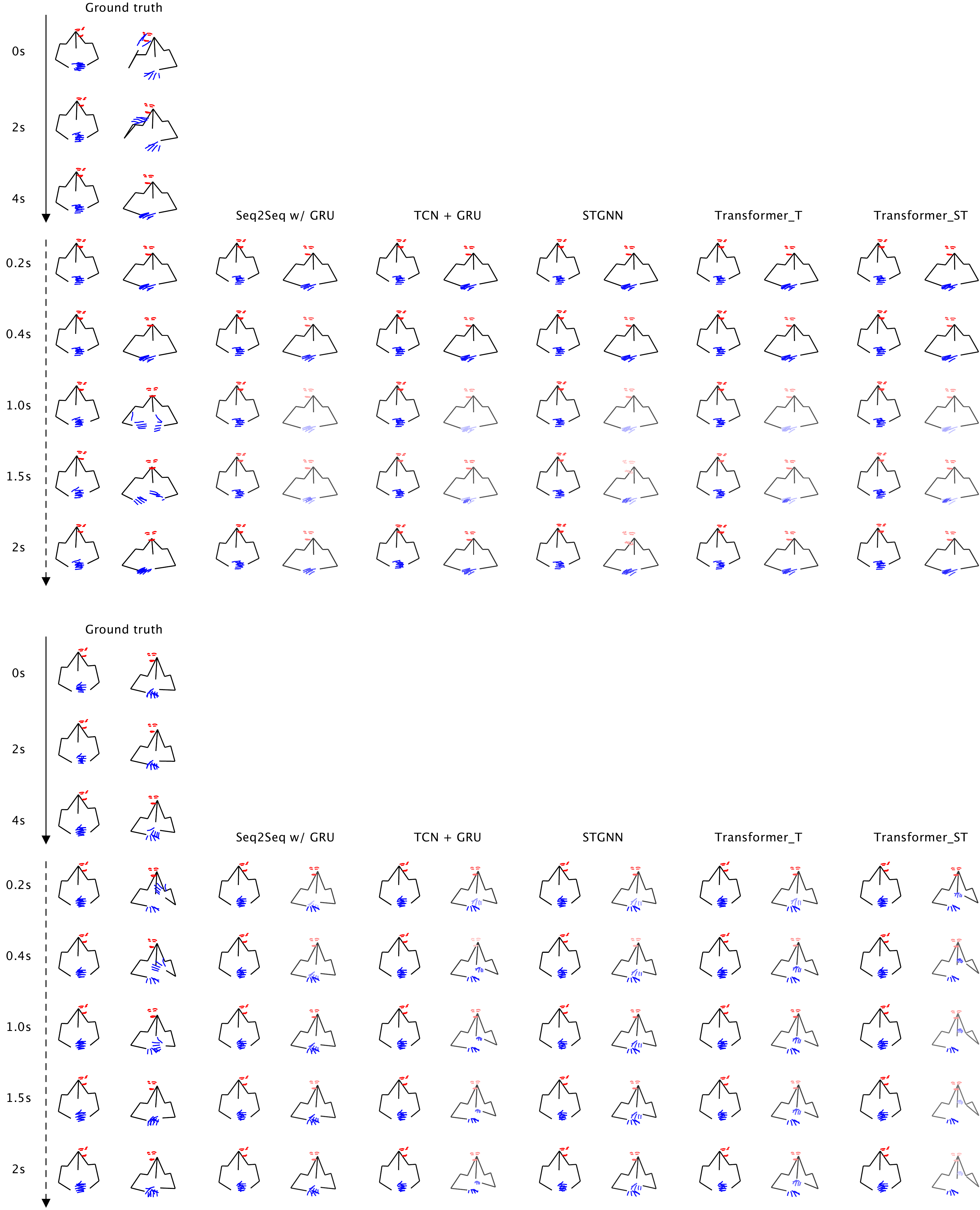}
    \caption{Two examples where remaining static represents a safer option than trying to replicate the emphasizing movement. In fact, at the bottom example, the freezing effect observed in Seq2Seq leads to a better overall accuracy than the riskier behavior elicited by the TCN and Transformer models.}
    \label{fig:visualization2}
\end{figure}

From the qualitative perspective, we observed that the predictions of our models tend to avoid taking risks. For example, they predict a sequence of fixed skeletons if the person has been static during the observation window, or tend to converge to the resting pose if the motion in the last observed frames is high. On the other side, some predictions feature emphasizing hand gestures, or head rotations towards the other interlocutor. In \autoref{fig:visualization} and \autoref{fig:visualization2}, we present four qualitative examples of diverse behaviors elicited by the short-term models. The alpha channel of the skeletons colors illustrates the error for each part of the body at each timestep (the more transparent, the higher the error). We can observe how transformer models successfully anticipate the motion of the left hand of the right-most participant in both examples from \autoref{fig:visualization}. \autoref{fig:visualization2} shows two scenarios where the difficulty of predicting the emphasizing hands gestures makes remaining static the safest bet. In fact, the transformer methods are penalized by their riskier predictions.

Given the superior performance of the short-term strategy, we chose it as our default strategy for all the experiments presented onward.

\subsection{Holistic experiments}
\label{sec:exp_holistic}

\begin{table}[t!]
    \footnotesize
    \centering
    \begin{tabular}{l@{\hskip 7mm}ccccc@{\hskip 7mm}cccc}
    \toprule
    & MPJPE & ST & MT & LT & FDE & $\Delta$ & $\Delta_{ST}$ & $\Delta_{MT}$ & $\Delta_{LT}$ \\
\midrule
\multicolumn{10}{c}{Face metrics} \\
\midrule
Ground truth & - & - & - & - & - & 0.65 & 0.62 & 0.66 & 0.67 \\
Zero-velocity & 13.29 & 4.35 & 11.99 & 17.65 & 18.80 & 0.00 & 0.00 & 0.00 & 0.00 \\
Transformer\_T & 12.77 & 3.74 & 11.33 & \textbf{17.25} & 19.08 & 0.12 & 0.21 & 0.09 & 0.10 \\
Transformer\_T $\mid$ Whole2Face & - & 3.55 & - & - & - & - & 0.25 & - & - \\
Transformer\_T $\mid$ Face2Face & \textbf{12.69} & \textbf{3.46} & \textbf{11.10} & 17.34 & \textbf{19.03} & 0.18 & 0.23 & 0.15 & 0.17 \\
\midrule
\multicolumn{10}{c}{Hands metrics} \\
\midrule
Ground truth & - & - & - & - & - & 1.33 & 1.28 & 1.31 & 1.36 \\
Zero-velocity & 25.87 & 9.04 & 20.37 & 31.98 & 34.51 & 0.00 & 0.00 & 0.00 & 0.00 \\
Transformer\_T & 25.13 & 8.37 & 19.78 & 31.14 & 33.43 & 0.15 & 0.45 & 0.09 & 0.08 \\
Transformer\_T $\mid$ Whole2Hands & - & 8.46 & - & - & - & - & 0.42 & - & - \\
Transformer\_T $\mid$ Hands2Hands & \textbf{24.87} & \textbf{8.07} & \textbf{19.62} & \textbf{30.84} & \textbf{33.31} & 0.14 & 0.37 & 0.11 & 0.07 \\
 \bottomrule
    \end{tabular}
    \caption{ Temporal transformer trained to either use the whole skeleton to predict the whole skeleton (Transformer\_T), the whole skeleton to predict only the face/hands (Whole2Face/Whole2Hands), or the face/hands to predict only the face/hands (Face2Face/Hands2Hands). We observe how target-specific models outperform the Transformer\_T model in almost all metrics.}
    \label{tab:holistic_comparison}
\end{table}

A question that quickly arises when predicting three different body parts is whether training a model to predict the behavior of each part would work better than training a generic model to predict all of them at once. To answer this question, we trained instances of the transformer model that only predict the motion of a single part of the body (face or hands). Each of them was trained twice with 1) the landmarks from the whole body (Whole2Face and Whole2Hands), 2) all landmarks except those from the part of the body to be predicted masked to zeros (Face2Face and Hands2Hands). Results for face and hands are summarized in \autoref{tab:holistic_comparison}. 
The Whole2Face and Whole2Hands could not be tested for the mid- and long-term windows because their design impeded to make recurrent inference.

In general, we observe a superior performance of the part-specific models. We hypothesize that one reason could be that the different levels of modeling complexity for each part of the body might lead to unequal training times needed to converge to an optimal solution. For instance, while training the Transformer\_T model, early-stopping might be triggered before getting to the actual lowest validation minimum for face if the hands prediction starts to overfit before.

\subsection{Dyadic experiments}
\label{sec:exp_dyadic}

\begin{table}[t!]
    \footnotesize
    \centering
    \begin{tabular}{lc@{\hskip 3mm}ccccc@{\hskip 6mm}cccc}
    \toprule
    & Params. & MPJPE & ST & MT & LT & FDE & $\Delta$ & $\Delta_{ST}$ & $\Delta_{MT}$ & $\Delta_{LT}$ \\
    \midrule
    Zero-velocity & - & 16.16 & 6.00 & 14.45 & 21.12 & 23.37 & 0.00 & 0.00 & 0.00 & 0.00 \\
\midrule
\multicolumn{10}{l}{Seq2Seq w/ GRU} \\
\midrule
Monadic & 12.0M & 15.96 & 5.73 & 14.36 & 20.88 & 23.18 & 0.09 & 0.21 & 0.06 & 0.06 \\
Dyadic - Early & 13.6M & 16.07 & 5.75 & 14.46 & 21.03 & 23.36 & 0.10 & 0.22 & 0.08 & 0.07 \\
Dyadic - Late & 15.2M & 15.92 & 5.76 & 14.38 & 20.77 & \textbf{22.82} & 0.07 & 0.18 & 0.05 & 0.04 \\
Dyadic - Interactive & 18.3M & 16.21 & 5.80 & 14.61 & 21.20 & 23.35 & 0.09 & 0.20 & 0.07 & 0.06 \\
\midrule
\multicolumn{10}{l}{TCN + GRU} \\
\midrule
Monadic & 1.7M & 16.28 & 5.45 & 14.07 & 21.79 & 25.59 & 0.17 & 0.25 & 0.09 & 0.20 \\
Dyadic - Early & 1.9M & 15.99 & 5.59 & 14.18 & 21.10 & 23.83 & 0.13 & 0.22 & 0.08 & 0.12 \\
Dyadic - Late & 1.7M & 16.32 & 5.48 & 14.10 & 21.85 & 25.61 & 0.17 & 0.23 & 0.09 & 0.20 \\
Dyadic - Interactive & 1.9M & 15.82 & 5.54 & 14.09 & 20.84 & 23.32 & 0.09 & 0.25 & 0.06 & 0.05 \\
\midrule
\multicolumn{10}{l}{Transformer\_T} \\
\midrule
Monadic & 15.3M & \textbf{15.65} & \textbf{5.44} & \textbf{13.93} & \textbf{20.63} & 23.06 & 0.12 & 0.29 & 0.08 & 0.08 \\
Dyadic - Early & 15.6M & 17.65 & 5.57 & 14.94 & 23.96 & 27.31 & 0.21 & 0.28 & 0.23 & 0.18 \\
Dyadic - Late & 16.6M & 16.48 & 5.55 & 14.50 & 21.89 & 24.39 & 0.14 & 0.25 & 0.15 & 0.10 \\
Dyadic - Interactive & 16.9M & 15.91 & 5.61 & 14.19 & 20.93 & 23.31 & 0.12 & 0.28 & 0.10 & 0.08 \\
 \bottomrule
    \end{tabular}
    \caption{Results of the dyadic architectures tested with three different fusion strategies. Monadic results correspond to those presented in \autoref{tab:unimodal_comparison}. We observe that the Transformer\_T approach still yields the best results.}
    \label{tab:dyadic_comparison_ST}
\end{table}

We adapted the Seq2Seq, TCN, and Transformer architectures such that they fuse the observed skeletons of both participants in three different ways. We left the STGNN adaptation for future work due to its higher conceptual complexity (e.g., design of dyadic graph structure). For all fusion strategies, the recurrency during inference contemplated both participants' predictions simultaneously. Unfortunately, results are inconclusive, as seen in Table~\ref{tab:dyadic_comparison_ST}. All fusion strategies performed slightly worse in the short term than the method considering only one participant (monadic). For the mid- and long-term windows, the results varied inconsistently across models. Interestingly, the training and validation curves showed a consistently faster convergence to the overfitting point for all dyad-aware methods. Although further research needs to be done, it seems that the higher dimensionality of the dyads requires much more data in order to generalize in the test set.

\subsection{Multimodalities}
\label{sec:exp_multimodal}

\begin{table}[t!]
    \footnotesize
    \centering
    \begin{tabular}{lc@{\hskip 2mm}ccccc@{\hskip 5mm}cccc}
    \toprule
    & Params. & MPJPE & ST & MT & LT & FDE & $\Delta$ & $\Delta_{ST}$ & $\Delta_{MT}$ & $\Delta_{LT}$ \\
    \midrule
Zero-velocity & - & 16.16 & 6.00 & 14.45 & 21.12 & 23.37 & 0.00 & 0.00 & 0.00 & 0.00 \\
Transformer\_T (Unimodal) & 15.3M & \textbf{15.65} & 5.44 & \textbf{13.93} & \textbf{20.63} & \textbf{23.06} & 0.12 & 0.29 & 0.08 & 0.08 \\
+ Metadata & 15.4M & 17.58 & 5.44 & 15.04 & 23.81 & 26.97 & 0.23 & 0.29 & 0.27 & 0.18 \\
+ Metadata w/ Personality & 15.4M & 16.55 & \textbf{5.42} & 14.57 & 22.05 & 24.69 & 0.18 & 0.29 & 0.20 & 0.13 \\
+ Audio & 15.4M & - & 5.46 & - & - & - & - & 0.30 & - & - \\
+ Transcriptions & 15.5M & 16.01 & 5.44 & 14.13 & 21.23 & 23.38 & 0.14 & 0.30 & 0.13 & 0.09 \\
 \bottomrule
    \end{tabular}
    \caption{Results of the Transformer-based bimodal architectures tested in this work (landmarks and another modality). Although there are no important differences in the short term, the architectures leveraging metadata seem to compromise their long-term performance. The audio model could not be recurrently tested for the long-term given the need of audio features specific to each observed frame.}
    \label{tab:multimodal_comparison_ST}
\end{table}

The Transformer\_T was trained by naively fusing several modalities of data: metadata, audio, and transcriptions. Results are summarized in Table~\ref{tab:multimodal_comparison_ST}. Both versions exploiting the audio and transcription modalities slightly underperform the unimodal version. Although the models incorporating metadata with and without personality perform apparently at the level of the unimodal approach in the short term, they predict much less accurate skeleton sequences for the long term. The relatively small number of participants (134), and metadata samples, might not be enough to generalize across participants. However, further research on more complex multimodal fusion strategies and on incremental combinations of multiple modalities is required. Similarly to the dyadic experiment, we also observed a faster decrease of the training loss and eventual divergence of the validation loss. The consistency of these observations reinforce our hypothesis that more data is needed in order to effectively model all behavioral combinations within the multimodal space.

\subsection{Additional experiments}
\label{sec:additional_exps}

\subsubsection{Robustness test}
\label{sec:exp_robustness_noise}

\begin{table}[t!]
    \footnotesize
    \centering
    \begin{tabular}{lc@{\hskip 4mm}ccccc@{\hskip 6mm}cccc}
    \toprule
    & MPJPE & ST & MT & LT & FDE & $\Delta$ & $\Delta_{ST}$ & $\Delta_{MT}$ & $\Delta_{LT}$ \\
    \midrule
Zero-velocity & \textbf{16.16} & 6.00 & \textbf{14.45} & 21.12 & 23.37 & 0.00 & 0.00 & 0.00 & 0.00 \\
Seq2Seq w/ LSTM & 16.85 & 6.26 & 15.55 & 21.73 & 23.88 & 0.10 & 0.29 & 0.07 & 0.04 \\
Seq2Seq w/ GRU & 16.82 & 6.18 & 15.35 & 21.81 & 24.19 & 0.12 & 0.30 & 0.08 & 0.07 \\
TCN + GRU & 16.81 & 5.93 & 14.67 & 22.30 & 26.00 & 0.18 & 0.30 & 0.09 & 0.20 \\
TCN + LSTM & 16.58 & 5.93 & 14.66 & 21.86 & 25.12 & 0.15 & 0.28 & 0.07 & 0.16 \\
STGNN & 17.67 & 6.10 & 15.38 & 23.53 & 26.52 & 0.18 & 0.32 & 0.17 & 0.14 \\
Transformer\_T & 17.12 & 6.65 & 15.64 & 22.06 & 24.40 & 0.15 & 0.41 & 0.09 & 0.08 \\
Transformer\_ST & \textbf{16.16} & \textbf{5.81} & 14.58 & \textbf{21.10} & \textbf{23.18} & 0.10 & 0.31 & 0.07 & 0.04 \\
 \bottomrule
    \end{tabular}
    \caption{Results of the short-term models when the inference is done with the raw annotations from UDIVA v0.5. These annotations include up to an extra 10\% of noise in the hands landmarks. Despite the significant increase in observed noise, the Transformer\_ST still outperforms the zero-velocity baseline, with a fair margin in the short term.}
    \label{tab:noise_comparison_ST}
\end{table}

As explained in Section~\ref{sec:evaluation_metrics}, our experiments were carried out with the \textit{cleaned} annotations of the UDIVA v0.5 test set. In a real deployment though, our model would necessarily leverage noisy annotations. In order to depict the expected decrease of performance in such conditions, we re-evaluated our models by using noisy annotations in the observation window instead (experiment inspired by \citealt{Corona2020}). More specifically, such noisy annotations correspond to the \textit{raw} test set from UDIVA v0.5. For fairness, note that the last observed skeleton was always preserved as cleaned in order to apply the predicted motion offsets on an initial correct skeleton. As the only differences between the \textit{cleaned} and the \textit{raw} annotations were the hands interpolation and the visibility fix~\citep{Palmero2022}, the noisy experiment was done with up to an extra 10\% of noisy hands annotations. Results are summarized in \autoref{tab:noise_comparison_ST}.

The results show a generalized drop in performance for all the models, especially for the Transformer\_T and the Seq2Seq models. It looks like the long-range modelling of TCNs and the spatial modeling of Transformer\_ST and STGNN increase their robustness of the models against noise.

\subsubsection{Attention analysis}
\label{sec:exp_attention}

\begin{figure}[t!]
    \centering
    \includegraphics[width=\textwidth]{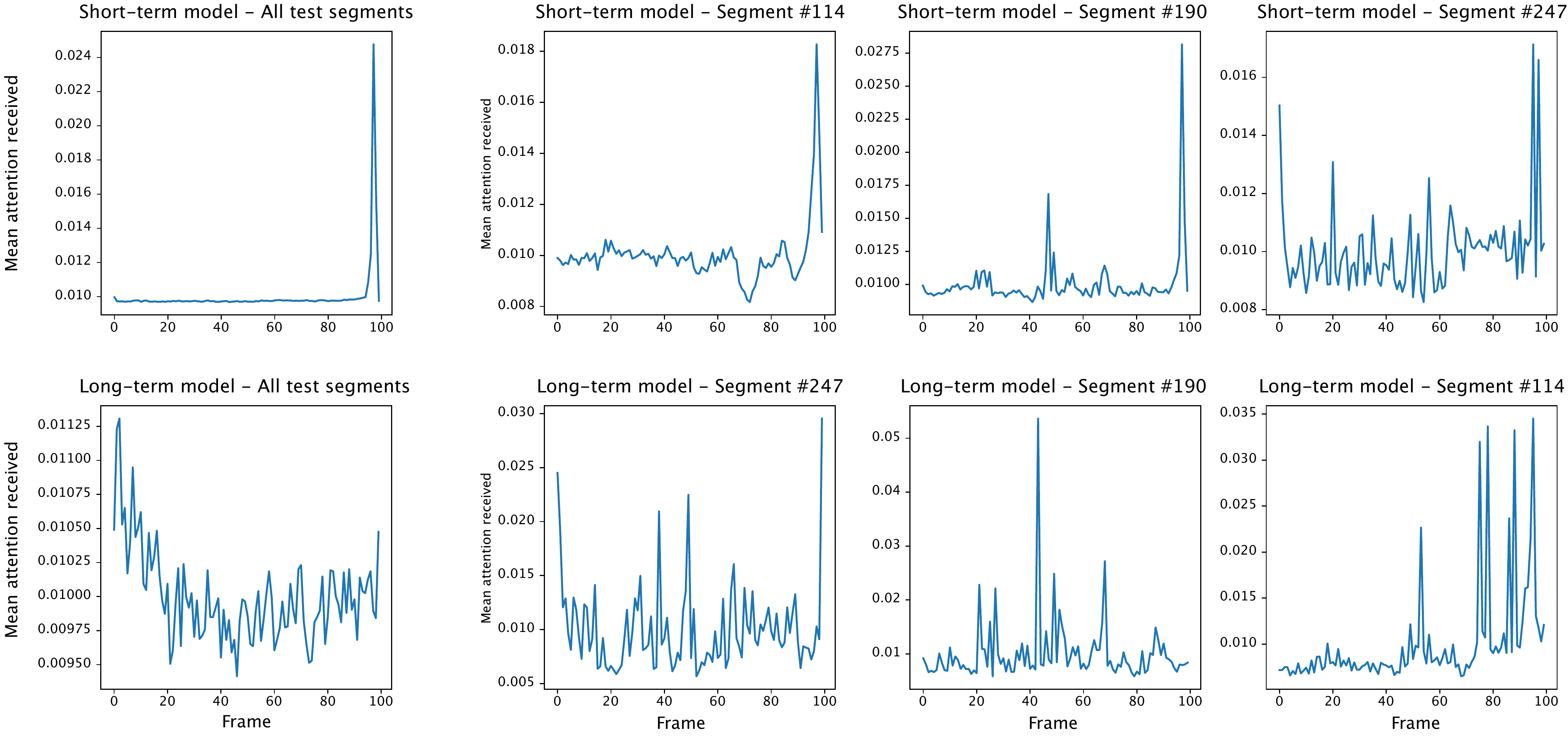}
    \caption{Attention given to each frame of the observation window during the encoding phase with the long- and short-term Transformer\_T models (top and bottom, respectively). The attention values across heads and transformer depths were averaged. A higher value intuitively means higher influence on the prediction. On the left, average attention values across all segments. On the right, attention values for several samples from the test set. We observe that, for the short-term models (top row), the information from the few latest observed frames were the most informative with regards to future behavior forecasting. This effect is not visible at all for the long-term model, which might try to extract mid- and long-range dependencies from the past.}
    \label{fig:attention}
\end{figure}

In general, human behavior is modeled by both short- and long-range temporal dependencies. During our experiments, we assumed that our models learnt to extract them in order to predict the future. However, we cannot be confident of such a statement. In fact, we do not know up to which extent these cues are useful for forecasting short- and long-term behavior. In this experiment, we try to determine this extent. To do so, we took advantage of the attention mechanism of the transformer-based model (Transformer\_T) and computed the average of the attention received by each frame across all the transformer depths and attention heads. Intuitively, this measure gives us a rough approximation of the importance attributed to each part of the observation window when building the encoding for the observation window. As a result, we can detect the most informative frames when it comes to predicting the future behavior. Results for the long- and short-term models are shown in \autoref{fig:attention} (top and bottom, respectively). We clearly observe a huge peak of attention in the last frames of the observation window when predicting the short-term future. Differently, for the long-term, the network may try to look for behavioral cues in the farther past. Of course, the last observed frames are less informative for the long term than for the short term.

After seeing the results of this experiment, an important question that raises is whether a model trained with only the last observed frames would predict as much accurate behavior as the model trained with the whole observation window (4s). To answer this question, we trained three short- and long-term models with much shorter observation windows (400, 200, and 80ms), see~\autoref{tab:obs_length_comparison}. For both models, the best overall results are yielded by the model trained with the whole observation window. However, these differences are smaller than expected, especially for the short term future, where the short-term model trained with only 400ms of observed behavior even outperforms the original model (MPJPE from 5.44 to 5.41). Surprisingly, the model considering only the last 80ms is fairly accurate in the short term. In return, the short-term models trained with shorter observation window perform worse in the long term. This suggests the importance of a long enough observation window, and leaves a more thorough exploration on its appropriate size as future work.

\begin{table}[t!]
    \footnotesize
    \centering
    \begin{tabular}{lccccc@{\hskip 6mm}cccc}
    \toprule
    & MPJPE & ST & MT & LT & FDE & $\Delta$ & $\Delta_{ST}$ & $\Delta_{MT}$ & $\Delta_{LT}$ \\
\midrule
\multicolumn{10}{c}{Long-term models} \\
\midrule
Transformer\_T - 4s & \textbf{17.27} & 6.24 & 15.30 & \textbf{22.72} & \textbf{25.38} & 0.17 & 0.25 & 0.17 & 0.13 \\
Transformer\_T - 400ms & 17.30 & \textbf{6.03} & \textbf{15.21} & 22.92 & 25.53 & 0.16 & 0.26 & 0.17 & 0.11 \\
Transformer\_T - 200ms & 17.80 & 6.04 & 15.61 & 23.69 & 26.34 & 0.19 & 0.30 & 0.20 & 0.14 \\
Transformer\_T - 80ms & 17.51 & 6.16 & 15.40 & 23.17 & 25.90 & 0.17 & 0.28 & 0.18 & 0.12 \\
\midrule
\multicolumn{10}{c}{Short-term models} \\
\midrule
Transformer\_T - 4s & \textbf{15.65} & 5.44 & \textbf{13.93} & \textbf{20.63} & \textbf{23.06} & 0.12 & 0.29 & 0.08 & 0.08 \\
Transformer\_T - 400ms & 15.83 & \textbf{5.41} & 14.10 & 20.89 & 23.21 & 0.10 & 0.27 & 0.06 & 0.06 \\
Transformer\_T - 200ms & 16.43 & 5.49 & 14.42 & 21.88 & 24.71 & 0.16 & 0.29 & 0.14 & 0.13 \\
Transformer\_T - 80ms & 16.37 & 5.55 & 14.48 & 21.69 & 24.19 & 0.15 & 0.31 & 0.13 & 0.10 \\

 \bottomrule
    \end{tabular}
    \caption{Performance of long- and short-term Transformer\_T models trained with smaller observation windows (400ms, 200ms, 80ms). Although leveraging 4s of observations yields the overall lowest error, short-term models trained with short observation lengths perform incredibly well, especially in the short term.}
    \label{tab:obs_length_comparison}
\end{table}

\subsubsection{Challenge results} 
\label{sec:exp_challenge}
\label{sec:challenge}

\begin{figure}[!t]
    \centering
    \includegraphics[width=\textwidth]{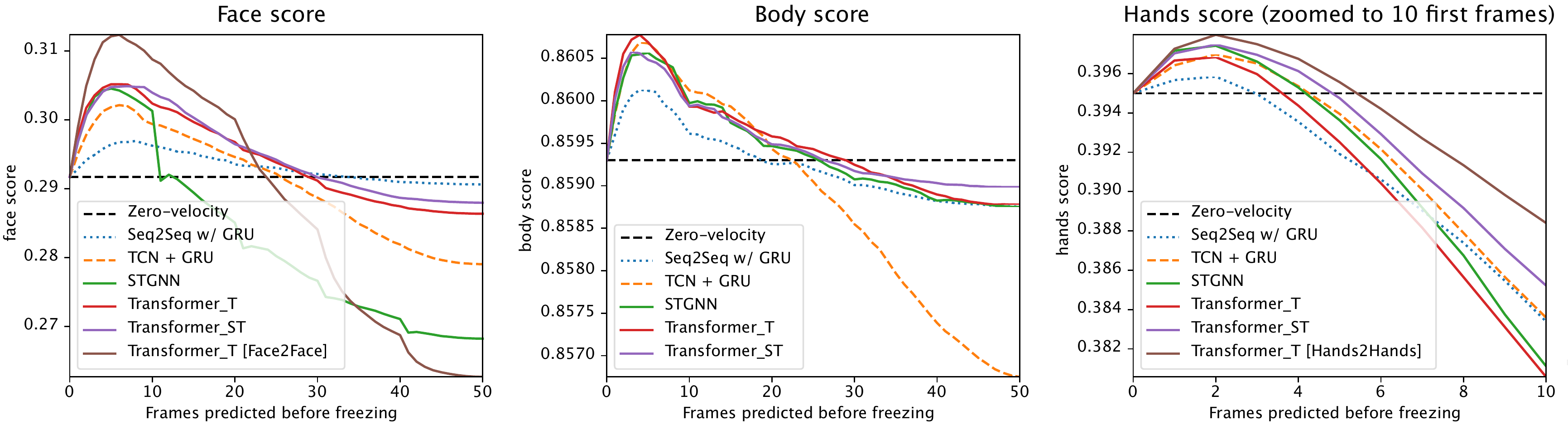}
    \caption{Performance of the proposed models in the validation stage of the DYAD'21 challenge split by face (left), body (middle) and hands (right). The \textit{x} axis corresponds to the number of frames predicted before starting to propagate the last predicted pose into the future (in a similar way the zero-velocity baseline is generated). For a better visualization, the hands plot is zoomed. The maximum is reached after predicting few frames, and quickly decreases as we dare to predict further due to the highly penalizing challenge metrics.}
    \label{fig:challenge_results}
\end{figure}

\begin{table}[t!]
    \footnotesize
    \centering
    \begin{tabular}{lccccc@{\hskip 6mm}cccc}
    \toprule
    Architecture & Face & Body & Hands \\
    \midrule
    Baseline (zero-velocity)     & 0.3458 & 0.8897 & 0.5392 \\
    rays2pix     & 0.3458 & 0.8824 & 0.5177 \\
    \citealt{tuyen2021forecasting}    & 0.2049 & 0.8507 & 0.3160 \\
    \midrule
    Transformer\_T (6-4-2*) &   0.3537 & \textbf{0.8898} & 0.5401 \\
    Transformer\_T $\mid$ Face2Face (6-0-0*) &  \textbf{0.3603} & - & - \\
    Transformer\_T $\mid$ Hands2Hands (0-0-2*)  & - & - & \textbf{0.5411} \\
    \midrule
    Ensemble of models    & 0.3603 & 0.8898 & 0.5411 \\
    \bottomrule
    \end{tabular}
    \vspace{2mm}
    \caption{Final results of our best model on the test set of the DYAD'21 challenge. The best model configurations from the validation stage were used as an ensemble of models to combine the benefits from each of them. Scores for the baseline (zero-velocity) and other participating teams are also included. *Number of frames predicted before freezing face, body and hands, respectively.}
    \label{tab:challenge_results}
\end{table}


In this section, we evaluate our architectures with the metrics of the ChaLearn LAP DYAD'21 behavior forecasting competition. The characteristics of the segments to predict were the same as the ones used in most of our experiments: an observation window of 100 frames (4 seconds), and a prediction window of 50 frames (2 seconds). However, we observed that accurately predicting behavior for the 2-seconds-long prediction window proposed in the challenge was beyond the capabilities of our networks. The main reason was that the challenge metrics greatly penalized wrong predictions, which were unavoidable given the stochasticity of the future. In order to compensate for those errors, the network would need to perfectly predict many segments, which was not feasible. Instead, we attempted a very conservative strategy. We knew from our experiments that our models made better predictions for the short-term future. Therefore, we explored whether by predicting fewer frames, $N$, and applying zero-velocity motion onward (propagating the last predicted pose into the future) we improved the zero-velocity baseline. \autoref{fig:challenge_results} illustrates the challenge metrics in function of the choice of $N$ for our best performing short-term models. We observe that this strategy indeed improves the baseline for very few frames ($<250$ms for face and $<200$ms for body and hands) before quickly falling as more frames are predicted before starting the static propagation. In order to make our final prediction at test stage, we first found the combination of model and $N$ value that maximized each metric (face, body, and hands). Then, we ensembled those models to generate the final predictions. Consistently with the results from the experiments from Section~\ref{sec:exp_holistic}, we found that the Face2Face and Hands2Hands models performed the best at predicting the face and hands behavior, respectively. With these two models and the Transformer\_T model for the body prediction, we scored 0.3603, 0.8898, 0.5411 for face, body, and hands (\autoref{tab:challenge_results}), respectively. The three scores are higher than those of the zero-velocity baseline and the participants from the DYAD'21 behavior forecasting competition.
\section{Discussion}
\label{sec:discussion}

In our multiple experiments, we came up with very interesting findings related to the size of the prediction window, or whether combining multiple modalities in a naive way improves the accuracy. In this section, we discuss all these points by encompassing them in generic areas.

\textbf{Long-term goal. }
From the unimodal experiments (Section~\ref{sec:exp_unimodal}) it can be observed that one needs to be conservative with respect to the prediction window length. The training of the long-term methods triggered the early stopping as soon as the network started predicting excessive risky motions, ending up with very static and freezing predictions. We hypothesize that this may be caused by two main factors. 
First, the high uncertainty present in such a long future time window favors the presence of many segments with similar observed behavior but different future outcomes. This eventually leads to a regression-to-the-mean effect. This is especially critical in rather static datasets like ours, in which the mean is remaining mostly immobile. In fact, the simple zero-velocity baseline yielded fairly accurate long-term results. 
Second, the complexity of learning to predict behavior for 2 seconds at once. Previous works~\citep{Wang2021Simple, Adeli2021} suggest the use of curriculum learning so that the network starts with easier sub-tasks (short-term) and gradually increases their difficulty (long-term). This strategy could improve our long-term networks' performance, and will be tested as future work. However, we believe that the first problem would still arise as soon as the network starts learning the long-term behaviors of the training set. 
We hypothesize that both the effective exploitation of extra information like multimodality, the scene context, or larger historical windows (history-aware methods), and the collection of bigger multimodal datasets will allow to reduce the future stochasticity, which will translate to more accurate behavior forecasting. 


\textbf{Short-term goal. }
In contrast with the subpar performance of the long-term models, the short-term models considerably outperform the zero-velocity baseline, even in the long term. In the first 400ms of prediction (short term), the error is similar for TCNs, STGNN, and Transformer methods. In particular, the spatio-temporal Transformer performs the best (MPJPE in the short term of 5.34px). The convolutional nature of TCNs and STGNN required to reduce their number of parameters in order to be trained in tractable time. Still, the training convergence took considerably longer than Seq2Seq and Transformer models. As a result, they show an incredibly good parameters/accuracy ratio. However, their high error and divergence in the long term suggests a compromised performance when applied recurrently. The Seq2Seq models, instead, offer a more modest improvement for the short-term future but deal very well with the recurrent future prediction. We have two hypotheses for this. First, their recurrent structure might learn more generalizing patterns to deal with the noise. Second, they may have better capabilities for deciding whether to predict static behavior (low divergence in the mid- and long-term futures). As a result, the good short-term accuracy would help favor a more accurate propagation of an almost frozen skeleton pose to the future. The qualitative results from \autoref{fig:visualization} illustrate this effect. In terms of accuracy, the Transformer approaches combine the benefits from both convolutional and recurrent worlds. Their excellent short-term performance also translates into a fairly good long-term accuracy. However, yet less severe, the freezing pattern is also observed in the Transformer-based architectures. Another interesting point is the incredibly low error increase of the spatio-temporal transformer when leveraging more noisy annotations (Section~\ref{sec:exp_robustness_noise}). We think that its joints-wise attention mechanism may be useful to filter noisy parts of the body out (e.g., hands). This hypothesis could be validated by analyzing the attention received by the most noisy landmarks, and remains as a very compelling future work. Also as future work, we will explore the benefits from the implementation of adversarial training. This would be particularly interesting because, given the massive amount of static segments in the UDIVA v0.5, generating realistic behavior does not necessarily mean generating moving sequences.


\textbf{Holistic. }
The holistic experiment showed a superior performance of the models trained to predict only a single part of the body. Our rationale behind this observation is that the validation minimum for each part of the body is reached at different training stages. As a result, the early-stopping of holistic models (Transformer\_T) may end up being triggered at the optimum point for hands. In fact, while the short-term face error for Whole2Face is lower than for Transformer\_T, the short-term hands error for the latter and Whole2Hands is similar. This suggests that the hands validation error could be the early-stopping trigger for the Transformer\_T model. Future work needs to explore more complex training strategies that address these issues. On the other hand, although the Whole2Hands and Whole2Face models should theoretically perform better than the Hands2Hands and Face2Face models, they do not. This may be caused by the lack of data available to learn all the behavioral dependencies among far away positioned joints. An alternative explanation is that the dependencies among parts of the body might not be useful enough for motion forecasting in the short-term future so that the network decides to put efforts into learning them. Additionally, we observed that the validation loss for face decreased steadier and took considerably longer than the hands' to overfit. This superior generalization capability observed for face may be due to the fact that the face mostly performs translation and rotations as a unit. Therefore, it can be easier modelled and predicted. The hand and the fingers, instead, can move more freely and may be harder to model and predict.



\textbf{Interlocutor-aware dynamics. }
None of our several attempts to merge the skeletons from both participants of the conversation improves the unimodal results. Our hypotheses behind it are mainly three. First, there is a natural imbalance between the interpersonal and intrapersonal dynamics in dyadic conversations~\citep{Ahuja2019}. As a result, the model might not find the minority class predictive enough to model it. In fact, recent methods that found improvements by modeling the interpersonal dynamics were evaluated in highly interactive scenarios like dancing or fencing~\citep{Honda2020, Katircioglu2021, Guo2021}. Secondly, the interpersonal dynamics are different for each pair of interacting subjects. Therefore, the data needed to model all possible interpersonal dependencies is considerably bigger. The steeper and faster decrease observed for the dyadic training losses supports this hypothesis. Finally, the approach used might have failed to capture interdependencies or jointly model both behaviors. More research is needed to come to any strong conclusion, as there are evidences that joint modeling is necessary and works~\citep{Ahuja2019, Hua2019}.


\textbf{Multimodalities exploitation. }
In this work, we trained three transformer-based models that naively merged the observed landmarks with either the audio, the transcriptions, or the metadata of the target person. All of them performed worse than the unimodal model. Similarly to the dyadic experiments, the multimodal training losses showed steeper decreases. Again, the extra information about the participant increases the dimensionality of the problem and thus the amount of data needed to acquire good generalization capabilities. For instance, the model with access to the participant's personality may learn to discern behavioral patterns specific to each type of personality. However, these learnt behavioral associations may be wrong due to the low amount of representations of each personality type in the training dataset (134 participants). The same argument can be extrapolated to audio and transcriptions. One more time, there is a need for larger and more diverse datasets in what refers to participants and elicited behaviors.

\textbf{Noise. } The noise present in the annotations from UDIVA v0.5 has a direct impact on our experiments from two perspectives. First, the faulty skeletons in the observation windows might interfere in the behavior modeling, which directly impacts the models' predictive capability. We noticed this effect in our robustness experiment (Section~\ref{sec:exp_robustness_noise}), where the increased noise from the raw observation windows considerably impacted the accuracy of the predicted behavior. Secondly, the noisy skeleton annotations used as future ground truth might be promoting conflicting gradient flows at training time, which may further accentuate the regression-to-the-mean effect. Nonetheless, most of our models still outperform the zero-velocity baseline for the short-term future, even when inference is run on noisy observations. Therefore, our work proves the effectiveness of weakly-supervised training by using noisy skeleton extractions, and its possibilities to a hypothetical real deployment by using automatically extracted annotations. Interestingly, the fairly big amount of datasets available with respect to contexts, tasks, and subjects may benefit from weakly- or even self-supervised learning and help towards the development of a definitive individual-agnostic behavioral model. Future work includes the exploration of specific techniques to better deal with the noise in the training data~\citep{ghosh2017learning}.

\textbf{Future work. } We foresee the evolution of future research towards several directions. First, we predict that the community interest will grow towards the exploration of behavior forecasting in the frequency space. Many works have already stated their benefits for avoiding the freezing behavior in single human motion forecasting~\citep{Mao2020, Mao2021}. In fact, few very recent works showed very promising results in this direction in highly dyad-driven scenarios like dancing~\citep{Katircioglu2021, Guo2021}. Second, we think that the biggest benefit will come from methods that are capable (at inference time) of adapting their learnt behavioral model to the specific behavioral patterns of the person for which forecasting is being applied. Meta-learning could be a good candidate to represent one of the next breakthroughs in this field~\citep{Raman2021, moon2021fast}. We also think that models that promote the interpretability of the behavioral predictions (e.g., causal models) may help us to automatically detect the dependencies among visual cues/signals, which may eventually lead to the discovery of new underlying mechanisms in human social communication. Finally, the exploitation of more explicit contextual cues, and the exploration of stochastic methods are still open issues.



\section{Ethics} 
\label{sec:ethics}

The research presented herein has several advantages and applications for good, but also comes with a number of potential pitfalls. Regarding the applications for good, social behavior forecasting is increasingly showing its potential due to the wide spectrum of possibilities it enables. Specifically, personalized interventions in virtual agents or assistive robots may improve well-being and mental health. For example, pedagogical agents~\citep{davis2018} which are personalized to maximize learner’s attention and learning are another important real-life application. Health care delivered by robots is also a burgeoning field of research~\citep{esterwood2021}, and personalization of robots’ personalities seem to have positive impacts of patients’ health and social outcomes~\citep{andrist2015} by means of increasing acceptance of robot’s care. Behavior forecasting can also have applications for good in assistive technology (e.g., anticipating falls), collaborative and autonomous robots or autonomous vehicles~\citep{Chaabane_2020_WACV}. However, each new technology comes with pitfalls and limitations. Non-consensual behavior forecasting may have potential pitfalls in areas such as security borders or migration controls, where unfair algorithms may lead to undesired outcomes~\citep{mckendrick2019} impacting human rights~\citep{ahkmetova2021}. All in all, although data protection regulations vary across countries~\citep{guzzo2015}, data privacy and data protection must ensure informational self-determination and consensual use of the information that can be extracted with the methods presented herein. In this sense, frameworks such as the EU General Data Protection Regulation (GDPR\footnote{\url{https://gdpr.eu/}.}) provide excellent safeguards for establishing ethical limits that should not be crossed.

\section{Conclusions}


In this work, we presented the first comprehensive comparison of state-of-the-art approaches for behavior forecasting during dyadic conversations. Differently from other human motion prediction works, our approach considered the participants' whole body (face, body, and hands landmarks). In particular, we proposed several adaptations of recurrent, convolutional, transformer-based, and graph networks for whole-body motion forecasting, most of them outperforming the zero-velocity baseline when trained under the right conditions (e.g., short-term optimization). Our experiments showed the superiority of the temporal and spatio-temporal transformer-based approaches. In fact, our temporal transformer represents the new state of the art in the ChaLearn LAP DYAD'21 competition. Interestingly, the model with spatio-temporal attention achieved better results than the baseline even when using the raw annotations. This proves the feasibility of weakly-supervised learning for human behavior forecasting tasks. Finally, our experiments helped us to identify, analyze and raise awareness of the main challenges posed by the UDIVA v0.5 dataset for behavior forecasting.

\acks{
Isabelle Guyon was supported by ANR Chair of Artificial Intelligence HUMANIA ANR-19-CHIA-0022. This work has been partially supported by the Spanish project PID2019-105093GB-I00 and by ICREA under the ICREA Academia programme.
}

\bibliography{jmlr-sample}

\appendix

\section{Architecture details}
\label{app:arch_details}

In all architectures tested, a first embedding layer transformed the input coordinates and offsets from all landmarks into an intermediate representation vector of size 512 by means of a dense layer. For bimodal architectures, the metadata, audio, and transcriptions representations were also embedded through a dense layer into representations of sizes 16, 64, and 64, respectively. All non-linearities used after convolutional or dense layer were leaky ReLUs with a negative slope of 0.01.

\subsection{Seq2Seq}
Both the encoder and the decoder consisted of one-layer LSTM or GRU units with hidden and cell states of size 1024. The hidden states from the encoder were used to initialize the decoder's. Two dense layers of 1024 units transformed the output from the decoder to the predicted skeleton pose. A residual layer was added from the input to the output of the decoder, so that this predicts the offsets of movement between future frames.

\subsection{TCN}
Causal dilated temporal convolutions without padding transformed the embeddings of the sequence of $N$ observed skeletons to a single vector whose receptive field was the whole observed window of skeletons. This was done by means of five sequential blocks of two temporal convolution: one with dilation factor $D$, and kernel size $K$ (no padding), and another with dilation factor 1 and kernel size 1 (no padding). The $K$ value was set to 2 for all blocks, and the $D$ was set to 1, 3, 9, 27, and 59 for the five blocks, respectively. The decoder consisted of a one-layer GRU or LSTM unit followed by two dense layers with 1024 and 512 units, respectively. The same residual layer as the one used in the Seq2Seq method was implemented.

\subsection{STGNN}

A total of three blocks were used. Each block consisted of a temporal layer and spatial layer. In the temporal layer, TCN with inception-like kernel set to 2, 3, 9, 11 was used. In the spatial layer, a mixhop graph neural network of maximum neighbour order 2 was used. The graph structure was parametrized by an embedding of dimension 40. A residual connection was used. A final skip convolution connecting each output of TCN, raw input, output was implemented. The TCN channel, residual channel, graph channel, and skip channel were of sizes 32, 32, 32, 64, respectively. A final MLP converted the intermediate vector of size 128 channels into the final output.

\subsection{Transformer}
For the spatio-temporal model, the backbone from \cite{poseformer2021} was used. The depth of the transformer was set to 4, the number of heads to 8, and the dropout value for the stochastic depth was set to 0.2. No dropout was applied for the attention mechanism. The output of the learnt weighted averaging layer was used as initial hidden state of a one-layered GRU decoder unit. The details of the GRU decoder are equal to those from the TCN model. For the temporal model, the spatial attention was removed.

\section{Unimodal results for face, body, and hands}
\label{app:unimodal_split}
In this section, we present the evaluation of the unimodal architectures tested in Section~\ref{sec:exp_unimodal} with metrics computed for each part of the body, see Tables~\ref{tab:st_face_results}, \ref{tab:st_body_results} and \ref{tab:st_hands_results}. We can observe a significant higher accuracy for face and body behavior forecasting than for hands, which move with considerably faster motion ($\Delta$=0.31/0.65/1.33 px/frame for body/face/hands, in average).

\begin{table}[t!]
    \footnotesize
    \centering
    \begin{tabular}{l@{\hskip 7mm}ccccc@{\hskip 7mm}cccc}
    \toprule
    & MPJPE & ST & MT & LT & FDE & $\Delta$ & $\Delta_{ST}$ & $\Delta_{MT}$ & $\Delta_{LT}$ \\
    \midrule
Ground truth & - & - & - & - & - &  0.65 & 0.62 & 0.66 & 0.67 \\
\midrule
Zero-velocity & 13.29 & 4.35 & 11.99 & 17.65 & 18.80 & 0.00 & 0.00 & 0.00 & 0.00 \\
LinearProp & 25.41 & 4.67 & 18.48 & 37.86 & 48.22 & 0.54 & 0.54 & 0.54 & 0.54 \\
RTOMean & 13.63 & 4.68 & 12.41 & 17.95 & 19.29 & 0.16 & 0.16 & 0.16 & 0.16 \\
RTOMean\_L & 13.58 & 4.67 & 12.37 & 17.86 & 19.15 & 0.17 & 0.17 & 0.17 & 0.17 \\
\midrule
Seq2Seq w/ LSTM & 13.05 & 4.22 & 11.87 & 17.30 & 18.43 & 0.05 & 0.10 & 0.04 & 0.04 \\
Seq2Seq w/ GRU & 12.85 & 4.14 & 11.68 & \textbf{17.04} & \textbf{18.19} & 0.06 & 0.11 & 0.04 & 0.05 \\
TCN + LSTM & 13.14 & 3.85 & 11.63 & 17.76 & 19.78 & 0.12 & 0.15 & 0.07 & 0.15 \\
TCN + GRU & 13.14 & 3.84 & 11.52 & 17.82 & 20.15 & 0.13 & 0.18 & 0.07 & 0.15 \\
TCN + MLP & 13.05 & 4.04 & 11.76 & 17.42 & 18.75 & 0.06 & 0.12 & 0.05 & 0.05 \\
STGNN & 12.75 & \textbf{3.65} & 11.42 & 17.19 & 18.65 & 0.14 & 0.21 & 0.13 & 0.12 \\
Transformer\_T & 12.77 & 3.74 & \textbf{11.33} & 17.25 & 19.08 & 0.12 & 0.21 & 0.09 & 0.10 \\
Transformer\_ST & \textbf{12.70} & 3.68 & 11.35 & 17.12 & 18.58 & 0.10 & 0.20 & 0.09 & 0.07 \\
 \bottomrule
    \end{tabular}
    \caption{Performance of the short-term models for face behavior forecasting.}
    \label{tab:st_face_results}
\end{table}

\begin{table}[t!]
    \footnotesize
    \centering
    \begin{tabular}{l@{\hskip 7mm}ccccc@{\hskip 7mm}cccc}
    \toprule
    & MPJPE & ST & MT & LT & FDE & $\Delta$ & $\Delta_{ST}$ & $\Delta_{MT}$ & $\Delta_{LT}$ \\
    \midrule
Ground truth & - & - & - & - & - & 0.31 & 0.29 & 0.32 & 0.31 \\
\midrule
Zero-velocity & 5.82 & 2.07 & 5.20 & 7.69 & 8.42 & 0.00 & 0.00 & 0.00 & 0.00 \\
LinearProp & 10.73 & 2.82 & 8.27 & 15.37 & 19.00 & 0.18 & 0.18 & 0.18 & 0.18 \\
RTOMean & 6.28 & 2.22 & 5.52 & 8.37 & 9.32 & 0.04 & 0.04 & 0.04 & 0.04 \\
RTOMean\_L & 6.41 & 2.30 & 5.61 & 8.54 & 9.52 & 0.08 & 0.08 & 0.08 & 0.08 \\
\midrule
Seq2Seq w/ LSTM & 5.83 & 2.04 & 5.23 & 7.72 & 8.41 & 0.02 & 0.06 & 0.01 & 0.01 \\
Seq2Seq w/ GRU & 5.80 & 1.99 & 5.17 & 7.70 & 8.41 & 0.03 & 0.06 & 0.02 & 0.02 \\
TCN + LSTM & 6.01 & 1.99 & 5.12 & 8.16 & 9.61 & 0.06 & 0.07 & 0.03 & 0.08 \\
TCN + GRU & 6.01 & \textbf{1.98} & 5.14 & 8.15 & 9.60 & 0.07 & 0.08 & 0.03 & 0.09 \\
TCN + MLP & 5.81 & 2.01 & 5.17 & 7.71 & 8.45 & 0.03 & 0.06 & 0.02 & 0.02 \\
STGNN & 5.85 & 1.99 & 5.16 & 7.80 & 8.64 & 0.04 & 0.07 & 0.03 & 0.04 \\
Transformer\_T & 5.80 & 2.02 & \textbf{5.11} & 7.72 & 8.55 & 0.04 & 0.09 & 0.03 & 0.03 \\
Transformer\_ST & \textbf{5.78} & 1.99 & 5.13 & \textbf{7.68} & \textbf{8.34} & 0.03 & 0.09 & 0.02 & 0.01 \\
 \bottomrule
    \end{tabular}
    \caption{Performance of the short-term models for upper body behavior forecasting.}
    \label{tab:st_body_results}
\end{table}

\begin{table}[t!]
    \footnotesize
    \centering
    \begin{tabular}{l@{\hskip 7mm}ccccc@{\hskip 7mm}cccc}
    \toprule
    & MPJPE & ST & MT & LT & FDE & $\Delta$ & $\Delta_{ST}$ & $\Delta_{MT}$ & $\Delta_{LT}$ \\
    \midrule
Ground truth & - & - & - & - & - & 1.33 & 1.28 & 1.31 & 1.36 \\
\midrule
    Zero-velocity & 25.87 & 9.04 & 20.37 & 31.98 & 34.51 & 0.00 & 0.00 & 0.00 & 0.00 \\
LinearProp & 55.00 & 12.20 & 38.94 & 77.91 & 97.87 & 0.95 & 0.95 & 0.95 & 0.95 \\
RTOMean & 28.81 & 9.78 & 22.49 & 36.29 & 39.32 & 0.26 & 0.26 & 0.26 & 0.26 \\
RTOMean\_L & 28.81 & 9.81 & 22.51 & 36.28 & 39.37 & 0.29 & 0.29 & 0.29 & 0.29 \\
\midrule
Seq2Seq w/ LSTM & 25.40 & 8.61 & 20.23 & 31.31 & 33.43 & 0.10 & 0.30 & 0.07 & 0.05 \\
Seq2Seq w/ GRU & 25.82 & 8.65 & 20.45 & 31.99 & 34.59 & 0.14 & 0.35 & 0.10 & 0.09 \\
TCN + LSTM & 25.83 & 8.45 & 19.97 & 32.36 & 36.32 & 0.18 & 0.37 & 0.08 & 0.18 \\
TCN + GRU & 26.31 & 8.28 & 19.95 & 33.36 & 38.21 & 0.24 & 0.38 & 0.13 & 0.27 \\
TCN + MLP & 25.69 & 8.53 & 20.18 & 31.95 & 34.60 & 0.13 & 0.32 & 0.08 & 0.09 \\
STGNN & 26.51 & 8.36 & 20.53 & 33.45 & 36.89 & 0.19 & 0.38 & 0.15 & 0.15 \\
Transformer\_T & \textbf{25.13} & 8.37 & \textbf{19.78} & \textbf{31.14} & 33.43 & 0.15 & 0.45 & 0.09 & 0.08 \\
Transformer\_ST & 25.15 & \textbf{8.16} & 19.81 & 31.24 & \textbf{33.15} & 0.12 & 0.40 & 0.08 & 0.04 \\
 \bottomrule
    \end{tabular}
    \caption{Performance of the short-term models for hands behavior forecasting.}
    \label{tab:st_hands_results}
\end{table}





\end{document}